%% file: 2026_Eger_Transforming_Science_with_LLMs.tex
\documentclass[manuscript,screen,review=false]{acmart}

\usepackage[breakable]{tcolorbox}
\definecolor{boxback}{HTML}{dde5ef}
\definecolor{boxframe}{HTML}{4b72a6}
\newcommand{\mybox}[1]{\begin{tcolorbox}[breakable, colback=boxback, colframe=boxframe, arc=2mm, boxrule=0.3mm, width=\textwidth]
#1
\end{tcolorbox}}

\usepackage{adjustbox}
\usepackage{smartdiagram}

\usepackage{multirow}
\usepackage{fontawesome5}

\usepackage{jabbrv}

\ifdefined\highlightchanges
\usepackage{mdframed}
\newcommand\changed[1]{\textcolor{blue}{#1}}
\newenvironment{changedpars}{\color{blue}}{}
\newenvironment{movedpars}{\color{green!66!black}}{}
\newenvironment{changedfloat}{%
\begin{mdframed}[backgroundcolor=blue!15,linewidth=0]
}{%
\end{mdframed}
}
\else
\newcommand\changed[1]{#1}
\newenvironment{changedpars}{}{}
\newenvironment{movedpars}{}{}
\newenvironment{changedfloat}{}{}
\fi

\setcopyright{rightsretained}
\copyrightyear{2026}
\acmYear{2026}
\acmDOI{XXXXXXX.XXXXXXX}

\begin{document}

\title[Transforming Science with Large Language Models]
{Transforming Science with Large Language Models: A Survey on AI-assisted Scientific Discovery, Experimentation, Content Generation, and Evaluation}


\author{Steffen Eger}
\email{steffen.eger@utn.de}
\orcid{0000-0003-4663-8336}
\affiliation{%
  \institution{University of Technology Nuremberg (UTN)}
  \city{Nuremberg}
  \country{Germany}
}

\author{Yong Cao}
\email{yong.cao@uni-tuebingen.de}
\orcid{0000-0002-3889-0382}
\affiliation{%
  \institution{University of Tübingen, Tübingen AI Center}
  \city{Tübingen}
  \country{Germany}
}

\author{Jennifer D'Souza}
\email{jennifer.dsouza@tib.eu}
\orcid{0000-0002-6616-9509}
\affiliation{%
  \institution{TIB Leibniz Information Centre for Science and Technology}
  \city{Hannover}
  \country{Germany}
}

\author{Andreas Geiger}
\email{a.geiger@uni-tuebingen.de}
\orcid{0000-0002-8151-3726}
\affiliation{%
  \institution{University of Tübingen, Tübingen AI Center}
  \city{Tübingen}
  \country{Germany}
}

\author{Christian Greisinger}
\email{christian.greisinger@utn.de}
\orcid{}
\affiliation{%
  \institution{University of Technology Nuremberg (UTN)}
  \city{Nuremberg}
  \country{Germany}
}

\author{Stephanie Gross}
\email{stephanie.gross@ofai.at}
\orcid{0000-0002-9947-9888}
\affiliation{%
  \institution{Austrian Research Institute for Artificial Intelligence}
  \city{Vienna}
  \country{Austria}
}

\author{Yufang Hou}
\email{yufang.hou@it-u.at}
\orcid{0000-0003-2897-6075}
\affiliation{%
  \institution{IT:U Interdisciplinary Transformation University Austria}
  \city{Linz}
  \country{Austria}
}

\author{Brigitte Krenn}
\email{brigitte.krenn@ofai.at}
\orcid{0000-0003-1938-4027}
\affiliation{%
  \institution{Austrian Research Institute for Artificial Intelligence}
  \city{Vienna}
  \country{Austria}
}

\author{Anne Lauscher}
\email{anne.lauscher@uni-hamburg.de}
\orcid{0000-0001-8590-9827}
\affiliation{%
  \institution{University of Hamburg}
  \city{Hamburg}
  \country{Germany}
}

\author{Yizhi Li}
\email{yizhi.li-2@manchester.ac.uk}
\orcid{0000-0002-3932-9706}
\affiliation{%
  \institution{University of Manchester}
  \city{Manchester}
  \country{United Kingdom}
}

\author{Chenghua Lin}
\email{chenghua.lin@manchester.ac.uk}
\orcid{0000-0003-3454-2468}
\affiliation{%
  \institution{University of Manchester}
  \city{Manchester}
  \country{United Kingdom}
}

\author{Nafise Sadat Moosavi}
\email{n.s.moosavi@sheffield.ac.uk}
\orcid{0000-0002-8332-307X}
\affiliation{%
  \institution{University of Sheffield}
  \city{Sheffield}
  \country{United Kingdom}
}

\author{Wei Zhao}
\email{wei.zhao@abdn.ac.uk}
\orcid{0000-0001-7249-0094}
\affiliation{%
  \institution{University of Aberdeen}
  \city{Aberdeen}
  \country{United Kingdom}
}

\author{Tristan Miller}
\email{Tristan.Miller@umanitoba.ca}
\orcid{0000-0002-0749-1100}
\affiliation{%
  \institution{University of Manitoba}
  \city{Winnipeg}
  \state{Manitoba}
  \country{Canada}
}

\renewcommand{\shortauthors}{Eger et al.}

\begin{abstract}
\changed{With the advent of large multimodal language models, science is now at a threshold of an AI-based technological transformation.  An emerging ecosystem of models and tools aims to support researchers throughout the scientific lifecycle, including (1)~searching for relevant literature, (2)~generating research ideas and conducting experiments, (3)~producing text-based content, (4)~creating multimodal artifacts such as figures and diagrams, and (5)~evaluating scientific work, as in peer review.  In this survey, we provide a curated overview of literature representative of the core techniques, evaluation practices, and emerging trends in AI-assisted scientific discovery.  Across the five tasks outlined above, we discuss datasets, methods, results, evaluation strategies, limitations, and ethical concerns, including risks to research integrity through the misuse of generative models.  We aim for this survey to serve both as an accessible, structured orientation for newcomers to the field, as well as a catalyst for new AI-based initiatives and their integration into future ``AI4Science'' systems.}
\end{abstract}



\maketitle

\input{introduction}
\input{methodology}

\section{AI Support for Individual Topics and Tasks}\label{sec:tasks}
\input{topics/literature}
\input{topics/experiments}
\input{topics/textgeneration}
\input{topics/multimodal}
\input{topics/peerreview}

\input{ethics}
\input{conclusion}

\begin{acks}
Yong Cao was supported by a VolkswagenStiftung Momentum grant. Jennifer D'Souza was supported by the \href{https://scinext-project.github.io/}{SCINEXT project} (BMBF, German Federal Ministry of Education and Research, Grant ID: 01lS22070). The NLLG Lab at UTN gratefully acknowledges support from the Federal Ministry of Education and Research (BMBF) via the research grant ``Metrics4NLG'' and the German Research Foundation (DFG) via the Heisenberg Grant EG 375/5-1. The work of Anne Lauscher is supported by the Excellence Strategy of the German Federal Government and the Federal States. Our AI use cases are documented in the supplemental material. 
\end{acks}
\bibliographystyle{ACM-Reference-Format-ISO4} 
\bibliography{2026_Eger_Transforming_Science_with_LLMs}

\clearpage
\input{appendix}

\end{document}

%% file: introduction.tex
\section{Introduction}
\label{sec:introduction}

Throughout history, science has undergone a number of paradigm shifts, culminating in today's era of data-intensive exploration~\cite{hey2009jim}. Although new tools and frameworks have accelerated the pace of scientific discovery, its basic steps have remained unchanged for centuries. These include (1)~conception of a research question or problem, typically arising from a gap in disseminated knowledge; (2)~collection and study of existing literature or data relevant to the problem; (3)~formulation of a falsifiable hypothesis; (4)~design and execution of experiments to test this hypothesis; (5)~analysis and interpretation of the resulting data; and (6)~reporting on the findings, allowing for their exploitation in real-world applications or as a source of knowledge for a further iteration of the scientific cycle. 

The advent of large multimodal foundation models, such as \href{https://chatgpt.com/}{ChatGPT}, \href{https://deepmind.google/technologies/gemini/}{Gemini}, \href{https://github.com/QwenLM/Qwen}{Qwen}, and \href{https://www.deepseek.com/}{DeepSeek}, is profoundly affecting many sectors of society, including scientific research.  Empirical evidence suggests that this influence extends well beyond computer science: an analysis of approximately 148,000 papers from 22 non-CS disciplines has revealed a rapid increase in citations of large language models (LLMs) between 2018 and 2024~\cite{pramanick2024transformingscholarlylandscapesinfluence}.  In parallel, a large global survey of researchers conducted by Wiley reported widespread expectations that use of AI will become mainstream in scientific practice in the next two years, despite its current use being often limited to writing assistance.\footnote{\url{https://www.wiley.com/en-us/ai-study}, \url{https://www.nature.com/articles/d41586-025-00343-5}}

While science has traditionally relied on human ingenuity and labor for generating research ideas, formulating hypotheses, searching for relevant literature, conducting experiments, and reporting results, recent AI systems have been promising support at every stage of this cycle. Examples include \href{https://elicit.com}{Elicit} and \href{https://ask.orkg.org/de}{ORKG ASK} for literature search, The AI Scientist~\cite{lu2024aiscientist} for experimentation, and AutomaTikZ~\cite{belouadi2024automatikz} and DeTikZify~\cite{belouadi2024detikzify} for multimodal content generation.  There is moreover growing interest in the use of AI for evaluating scientific outputs through automated peer review~\cite{10.1613/jair.1.12862}. \changed{Collectively, these advancements suggest the emergence of an integrated AI-assisted research workflow with the potential to accelerate discovery and streamline the documentation and communication of results.}\footnote{\changed{The benefits could be particularly significant for non-native speakers of English and those with lower technical skills, potentially increasing diversity and inclusivity in research.}}  The consolidation of a research community around AI-assisted science is evidenced by the establishment in 2024 and 2025 of dedicated venues such as the workshops on Natural Scientific Language Processing and Research Knowledge Graphs (NSLP)~\cite{rehm2024natural}, Foundation Models for Science (\href{https://fm-science.github.io/}{FM4Science}), AI \& Scientific Discovery (\href{https://ai-and-scientific-discovery.github.io/}{AISD}), Towards a Knowledge-grounded Scientific Research Lifecycle (\href{https://sites.google.com/view/ai4research2024}{AI4Research}), AI Agents for Science (\href{https://agents4science.stanford.edu/}{Agents4Science}), and Human--LLM Collaboration for Ethical and Responsible Science Production (SciProdLLM)~\cite{zhao2025first}.

\changed{Despite the rapid progress in this area, existing surveys typically focus on specific domains, such as applications in the social sciences or physics~\cite[e.g.,][]{XU2024103665,Zhang2023ArtificialIF}, or on a relatively narrow set of research tasks and ethical concerns~\cite[e.g.,][]{hastings2023ai,zhang-etal-2024-comprehensive-survey,luo2025llm4srsurveylargelanguage}.  To address this gap, the present survey adopts a workflow-centric perspective, providing a broad, cross-cutting overview of five central aspects of AI support for the research cycle: (1)~literature search and summarization (§\ref{sec:literature_search}); (2)~scientific experimentation and research idea generation (§\ref{sec:experiments}); (3)~unimodal generation and refinement of textual content, including titles, abstracts, and citations (§\ref{sec:textgeneration}); (4)~multimodal content generation and interpretation, including figures, tables, slides, and posters (§\ref{sec:multimodal}); and (5)~AI-assisted peer review (§\ref{sec:peer_review}).  
Rather than aiming for comprehensive coverage within each area, we focus on representative approaches that capture core methodological ideas and allow meaningful comparison across tasks.}

\changed{Ethical considerations are paramount in any discussion of AI in science.  Current tools exhibit a number of problems and limitations, including ``hallucination'', bias, limited reasoning abilities, and substantial environmental costs, and mechanisms for evaluating their output remain underdeveloped.  Broader concerns include the risks of ``fake science'', plagiarism, and erosion of research integrity through diminished human oversight.  Recent policy guidance on the use of AI in science, such as that of the EU,}\footnote{~\url{https://research-and-innovation.ec.europa.eu/document/download/2b6cf7e5-36ac-41cb-aab5-0d32050143dc_en?filename=ec_rtd_ai-guidelines.pdf}} \changed{emphasizes both the transformative potential of these technologies and the risks they pose if deployed without appropriate safeguards.  In this survey, these ethical considerations are addressed alongside our treatment of the appertaining research tasks, as well as in a dedicated discussion in §\ref{sec:ethics}.}


\begin{figure*}
  \begin{changedfloat}
  \centering
  \includegraphics[width=\textwidth]{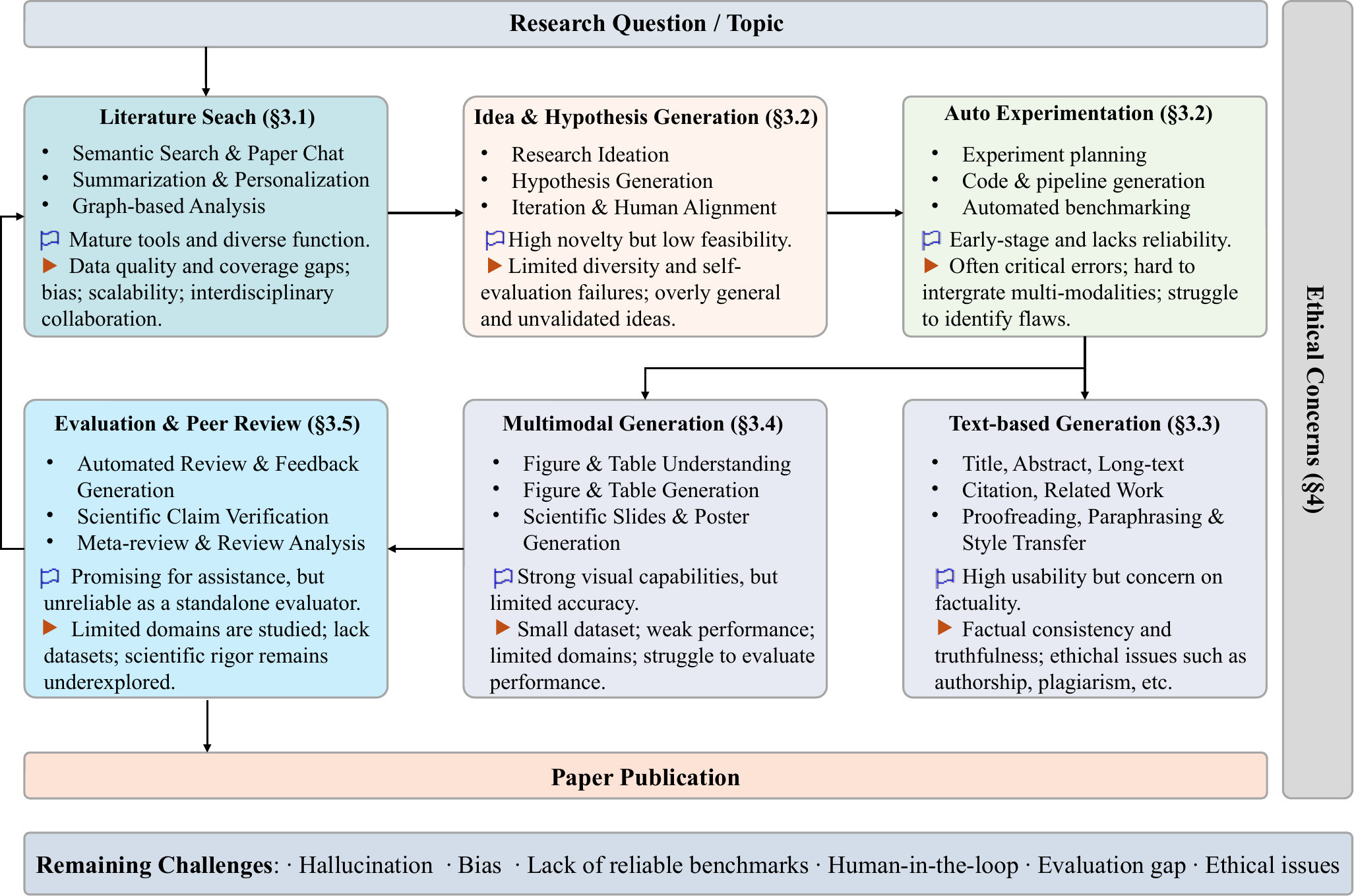}
  \caption{\changed{Overview of the AI-assisted scientific research workflow and remaining challenges, illustrating how AI support different stages of the research process. Each block summarizes the current status of AI capabilities (\textcolor{blue}{\faFlag[regular]}) and their limitations (\textcolor[HTML]{D45723}{$\blacktriangleright$}) at that stage.}}
  \label{fig:overview}
  \end{changedfloat}
\end{figure*}

\changed{As illustrated in Fig.~\ref{fig:overview}, the remainder of this paper is structured as follows. In §\ref{sec:methodology} we discuss the scope and methodological approach of our survey. The subsections of §\ref{sec:tasks} review representative literature for individual tasks in the research lifecycle, presenting datasets, methods, evaluation practices, limitations and future directions, and connections across tasks.  In §\ref{sec:ethics} we address broader ethical and integrity concerns.  In §\ref{sec:conclusion} we conclude with a synthesis of opportunities and challenges for AI-assisted scientific research.  The paper's appendix includes background material on the scientific discovery cycle, further elaboration on AI support for specific topics and tasks, and a list of abbreviations.  At \url{https://github.com/NL2G/TransformingScienceLLMs} we maintain a periodically updated list of further resources relating to this survey.}


%% file: methodology.tex
\section{Survey Scope and Methodology}
\label{sec:methodology}

\changed{This survey provides a broad, workflow-centric overview of AI methods and applications that support scientific research across the full research lifecycle. 
It 
is intended primarily for researchers in AI-related fields (e.g., natural language processing, computer vision, and machine learning) seeking a structured orientation to this rapidly evolving area, with clear entry points for deeper exploration. Some of the material will also be useful to policymakers, practitioners, and research collaborators in adjacent fields, including human--computer interaction, library and information science, communication studies, metascience, science journalism, and research ethics.}

\changed{Given our topic's wide scope, rapid progress, and dependence on knowledge and methods from different domains, we adopt a narrative rather than a systematic survey methodology.  This approach is particularly well suited to synthesizing heterogeneous and evolving bodies of work, enabling connections to be drawn across domains and methodological traditions~\cite{king2005understanding,byrne2016improving,pare2015synthesizing}. Rather than imposing rigid inclusion and exclusion criteria, the narrative approach allows the survey to emphasize conceptual coherence, methodological representativeness, and comparability across tasks. At the same time, it requires transparency about scope and limitations: the survey does not aim to be exhaustive, nor does it claim to capture every recent publication.}

\changed{The papers and tools discussed in each subsection were selected by diverse co-author teams with domain expertise, using a common set of guiding principles. Initial pools of candidate works were typically assembled by combining (i)~seed papers known to the co-authors for their technical depth or influence, expanded through forward and backward citation analysis, and (ii)~targeted keyword searches in major scholarly databases such as Google Scholar. From these pools, works were selected on the basis of their relevance to the task under discussion, the maturity and clarity of their methodology, the reputation of the publication venue, and indicators of impact such as citation patterns relative to publication date. Preference was given to approaches that exemplify core techniques, employ well-defined evaluation protocols, or have served as reference points for subsequent work.  This selection strategy was intended to foreground approaches that are representative of broader research trends, amenable to comparison, and likely to remain relevant at least in the near term, thereby supporting the survey's integrative goals. Throughout §\ref{sec:tasks}, individual subsections use a common structure to highlight common evaluation dimensions, methodological trade-offs, and connections between tasks; this helps to situate individual contributions within a larger picture of AI-assisted scientific research and reduce any single-viewpoint bias inherent in our selection process.}


%% file: topics/literature.tex
\subsection{Literature Search, Summarization, and Comparison}
\label{sec:literature_search}

\mybox{The rapid growth of scientific literature presents a significant challenge for researchers to efficiently search, analyze, and summarize information.  AI-powered tools are transforming these tasks by leveraging NLP, machine learning~(ML), LLMs, citation and knowledge graphs~(KGs) to automate the retrieval, extraction, and summarization of scientific information. This section surveys state-of-the-art AI-enhanced literature discovery tools, categorized according to their core functionality: (1)~\textbf{AI-enhanced search}, which retrieves relevant literature from vast repositories; (2)~\textbf{graph-based systems}, which map relationships between research concepts and publications; (3)~\textbf{paper chat and QA}, which enable interactive exploration of scientific content; and (4)~\textbf{recommender systems}, which suggest relevant papers based on user preferences. (We further discuss traditional search engines and benchmarks with leaderboards in Appendix~\ref{ax:search_engine}.)}

\subsubsection{Data}

Scientific search engines rely on vast publisher databases to provide access to scientific literature. Understanding the classification of these repositories is essential for assessing search engines' coverage, reliability, and effectiveness in evidence-based research. Repositories vary by \textbf{access model}, \textbf{subject focus}, and \textbf{content type}, each serving a distinct role in academic discovery and knowledge dissemination. By \textbf{access model}, repositories fall into \emph{open access repositories}, which provide unrestricted access to research articles (e.g., \href{https://pubmed.ncbi.nlm.nih.gov/}{PubMed Central}, \href{https://arxiv.org/}{arXiv}); \emph{subscription-based repositories}, requiring institutional or individual subscriptions (e.g., \href{https://www.sciencedirect.com/}{ScienceDirect}, \href{https://link.springer.com/}{SpringerLink}); and \emph{hybrid repositories}, offering both free and paywalled content (e.g., \href{https://www.tandfonline.com/}{Taylor \& Francis Online}, \href{https://academic.oup.com/}{Oxford Academic}). By \textbf{subject focus}, repositories are either \emph{multidisciplinary}, covering broad disciplines (e.g., \href{https://clarivate.com/academia-government/scientific-and-academic-research/research-discovery-and-referencing/web-of-science/}{Web of Science}, \href{https://www.scopus.com/home.uri}{Scopus}), or \emph{subject-specific}, specializing in fields such as medicine (\href{https://pubmed.ncbi.nlm.nih.gov/}{PubMed}), physics (\href{https://inspirehep.net/}{INSPIRE-HEP}), and social sciences (\href{https://www.ssrn.com/index.cfm/en/}{SSRN}). By \textbf{content type}, \emph{institutional repositories} archive research outputs from specific organizations (e.g., \href{https://dspace.mit.edu/}{MIT DSpace}, \href{https://dash.harvard.edu/}{Harvard DASH}); \emph{preprint repositories} enable early dissemination of research before peer review (e.g., \href{https://www.biorxiv.org/}{bioRxiv}, \href{https://chemrxiv.org/engage/chemrxiv/public-dashboard}{chemRxiv}); and \emph{government and public sector repositories} provide access to publicly funded research (e.g., \href{https://ui.adsabs.harvard.edu/}{NASA ADS}, \href{https://www.openaire.eu/}{OpenAIRE}). \emph{Data repositories} (e.g., \href{https://datadryad.org/stash}{Dryad}, \href{https://zenodo.org/}{Zenodo}) store research datasets, supporting transparency and reproducibility, while \emph{aggregator repositories} (e.g., \href{https://www.base-search.net/}{BASE}, \href{https://core.ac.uk/}{CORE}) 
index content from multiple sources for broader searches. Lastly, \emph{grey literature repositories} (e.g., \href{https://opengrey.eu/}{OpenGrey}, \href{https://ethos.bl.uk/}{EThOS}) provide access to non-traditional research outputs such as theses, reports, and white papers, which may not be available through conventional publisher platforms.

The structure of scientific repositories shapes AI-enhanced search. While broad AI-based search engines like \href{https://elicit.com}{Elicit} and \href{https://ask.orkg.org}{ORKG ASK} query multiple publisher repositories, similar to \href{https://scholar.google.com}{Google Scholar}, tools like \href{https://notebooklm.google}{NotebookLM} focus on user-selected documents, and recommender systems such as \href{https://www.scholar-inbox.com/}{Scholar Inbox} rank new literature by relevance. AI-driven search enables customizable knowledge bases while optimizing discovery, retrieval, and personalization in research.

\subsubsection{Methods and Results} Here, we discuss four types of state-of-the-art AI-enhanced search tools.

\paragraph{AI-enhanced Search}

Platforms such as \href{https://elicit.com}{Elicit}, \href{https://consensus.app}{Consensus}, \href{https://openscholar.allen.ai}{OpenScholar}~\cite{openscholar}, and \href{https://typeset.io}{SciSpace} leverage AI, including LLMs, to extend beyond traditional search by enabling semantic search, paper summarization, evidence synthesis, and trend analysis. Unlike conventional search engines that rely on keyword matching, these tools use NLP and machine learning to extract key insights, synthesize information to answer research queries~\cite{giglou2024llms4synthesis}, and generate structured summaries~\cite{he-etal-2025-pasa, zhang-etal-2025-scientific, weng2025cycleresearcher}. Their ability to quickly summarize and categorize findings—such as study outcomes, methodologies, and limitations—helps researchers efficiently compare and interpret literature.

\input{table/literature_overview_table}

\paragraph{Graph-based Systems}

Graph-based systems such as \href{https://ask.orkg.org}{ORKG ASK}~\cite{oelen2025introducing} are designed to facilitate structured access to scientific knowledge. Unlike conventional paper search engines, they leverage a KG that organizes research contributions as structured data rather than unstructured text. Such contributions are typically extracted from the abstract, introduction, and result sections~\cite{dsouza-etal-2021-semeval, pramanick2024naturenlpanalyzingcontributions}. Those systems enable users to ask complex, domain-specific questions and receive answers synthesized from semantically structured scientific data. They typically use techniques such as KG-based reasoning and retrieval-augmented generation~(RAG) to extract relevant information from the KG, providing more interpretable and verifiable answers compared to traditional LLM-based QA systems. \href{https://citespace.podia.com/}{CiteSpace} and \href{https://sci2.cns.iu.edu/user/index.php}{Sci2} are specialized bibliometric analysis and network analysis tools to study the structure and evolution of scientific research. \href{https://citespace.podia.com/}{CiteSpace} focuses on identifying research trends, keyword co-occurrence networks, and citation bursts, using visual analytics to highlight emerging topics and influential papers using graphs. \href{https://sci2.cns.iu.edu/user/index.php}{Sci2} is a more general-purpose tool designed for analyzing scholarly datasets, enabling users to perform network analysis, geospatial mapping, and temporal modeling of scientific literature and collaboration patterns. \href{https://www.connectedpapers.com/}{\changed{Connected Papers}}
\changed{is a visual literature exploration tool that maps papers related to a seed paper to provide an overview of a research field and support tasks such as bibliography construction and identification of prior and derivative work. Instead of building a direct citation tree, it organizes papers using similarity scores based primarily on \emph{bibliographic coupling} and \emph{co-citation}, typically via normalized overlap-based measures~\cite{kessler1963bibcoupling,small1973cocitation}. The resulting weighted similarity graph visually clusters closely related papers and separates weaker ones, enabling interactive exploration of research clusters, and is powered by large-scale scholarly metadata (e.g., Semantic Scholar).}

\paragraph{Paper Chat and QA}

Paper chat and question-answering~(QA) systems such as \href{https://chatgpt.com/}{ChatGPT}, \href{https://chat.deepseek.com}{Deepseek Chat}, \href{https://notebooklm.google}{NotebookLM}, \href{https://www.explainpaper.com/}{ExplainPaper}, \href{https://www.chatpdf.com/}{ChatPDF}, and \href{https://docanalyzer.ai/}{DocAnalyzer.AI} allow users to interact with scientific papers by asking questions and receiving responses based on the document's content. They typically process a limited number of user-provided PDFs or text from specific websites. The core technology behind them is RAG~\cite{lewis2020retrieval, asai2024selfrag, kang-etal-2024-taxonomy}, a technique that combines information retrieval with LLMs
to improve accuracy and grounding. A typical RAG system first partitions the document into smaller sections and converts them into vector representations using embedding models. Upon a user query, the system retrieves the most relevant sections based on semantic similarity and passes them as context to an LLM, which then generates a response. This mechanism ensures that answers are directly grounded in the provided documents rather than relying solely on the model's pre-trained knowledge, enhancing factual reliability and interpretability. Some systems incorporate LLM agents~\cite{tan-etal-2023-multi2claim, cai-etal-2024-mixgr,li-etal-2025-chatcite} that can reason over retrieved information, summarize findings, or extract key insights. These agents can follow multi-step reasoning strategies to provide more nuanced responses, such as synthesizing information from multiple sections or explaining technical terms in simpler language. By anchoring responses to document content, RAG-based systems mitigate hallucinations and make it easier for users to verify claims by checking the referenced passages. The effectiveness of these systems depends on the quality of document chunking, the efficiency of retrieval, and the model's ability to integrate information into coherent, context-aware answers.

\paragraph{Recommender Systems} 

Scientific paper recommender systems such as \href{https://arxiv-sanity-lite.com/}{Arxiv Sanity}, \href{https://www.scholar-inbox.com/}{Scholar Inbox}, \href{https://researchtrend.ai}{ResearchTrend.ai}, and \href{https://www.researchrabbit.ai/}{Research Rabbit} leverage machine learning and information retrieval techniques to help researchers discover relevant literature. These systems generally fall into two main categories: content-based filtering, collaborative filtering and hybrid approaches. Content-based methods~\cite{amami2016lda, bhagavatula-etal-2018-content} analyze the text of papers to build representations that capture their meaning. Traditional approaches rely on sparse abstract or document representations such as TF--IDF~\cite{sparck1972statistical}, which assigns importance to words based on their frequency and distinctiveness in a corpus. More advanced models, such as SPECTER~\cite{cohan-etal-2020-specter} and GTE~\cite{li2023towards}, use dense abstract or document embeddings derived from neural networks; they map papers into a high-dimensional vector space where similar documents are close to each other. The Massive Text Embedding Benchmark (MTEB)~\cite{muennighoff-etal-2023-mteb} ranks many state-of-the-art embedding models on a comprehensive benchmark comprising various different datasets and tasks. These embeddings enable fast similarity searches and improve over simple keyword matching. In contrast, collaborative filtering~\cite{wang2014relational, bansal2016ask} relies on user interactions, such as downloads, bookmarks, and citations, to recommend papers based on the behavior of similar users. One challenge of pure collaborative filtering is the cold start problem, where new papers or users lack sufficient data for recommendations. To mitigate this, many modern systems employ hybrid approaches, such as two-tower architectures~\cite{10.1145/3298689.3346996, covington2016deep, yu2021dual}. These models learn separate representations for papers and users, combining textual embeddings with user interaction data to generate more personalized recommendations. State-of-the-art systems often use a mix of these techniques to balance relevance, novelty, and diversity. The effectiveness of these systems depends on the quality of embeddings, the availability of interaction data, and the efficiency of ranking algorithms that surface the most useful papers.

\begin{changedpars}
\subsubsection{Domains of Application}

The tools discussed in this section are largely domain-agnostic and can be applied across scientific disciplines by adapting the underlying corpus and domain resources. For example, in medical and neuroscience research, \citet{ezzdine2025physical} survey exercise and cognitive-training interventions for neurodegenerative disorders and discuss how AI methods are being used to support evidence synthesis and analysis. In ecology, LLMs have been evaluated for extracting structured ecological variables from the scientific literature to accelerate evidence synthesis~\cite{gougherty2024testing}. In chemistry and materials science, NLP\slash LLM pipelines have been used to mine synthesis conditions and material-property records from papers to construct structured datasets~\cite{zheng2023chatgpt,shetty2023polymer}. However, many benchmarks used to evaluate such systems remain concentrated in computer science and AI, reflecting current dataset availability. 
\end{changedpars}

\subsubsection{Limitations and Future Directions}

A primary challenge for scholarly search systems is \emph{data quality and coverage gaps}: systems often struggle with incomplete, non-standard, or outdated data sources, which can lead to inaccuracies and inconsistencies in retrieved information. There is also the issue of \emph{model bias}, where search and ranking algorithms adopt biases of their training data, potentially influencing the visibility of certain research areas and limiting the diversity of perspectives presented to users. Another major limitation lies in \emph{scalability and real-time processing}---i.e., efficiently handling large-scale datasets while maintaining low latency and high retrieval accuracy. 
\changed{Finally, many AI-assisted research tools rely on proprietary data, closed APIs, or evolving LLM backends, which complicates strict reproducibility and long-term comparability.}
These limitations suggest several promising future directions. One potential avenue is \emph{enhanced personalization}, which can be achieved by adapting search engines to user preferences, providing more tailored recommendations based on research interests and behavioral patterns. Fostering \emph{interdisciplinary collaboration} through the integration of AI-powered search systems with other digital tools, such as data visualization platforms and research management software, could likewise facilitate more comprehensive and insightful research outcomes. 


%% file: table/literature_overview_table.tex
\begin{table}[t]
\begin{changedfloat}
\caption{Overview of popular literature search, summarization, and comparison tools and their key features. \changed{$\checkmark$ indicates feature availability; empty cells indicate lack of features or publicly documented support.}}
\label{tab:google_scholar}
\vspace{1cm}
\centering
\resizebox{0.85\columnwidth}{!}{%
\begin{tabular}{lllll|llll|lllll|lllll|ll}
\multicolumn{1}{c}{\textbf{}} & \multicolumn{1}{c}{\textbf{Platform}} & \multicolumn{1}{c}{\rotatebox[origin=l]{45}{\makebox[0pt][l]{\textbf{Search}}}} & \multicolumn{1}{c}{\rotatebox[origin=l]{45}{\makebox[0pt][l]{\textbf{Recommendations}}}} & \multicolumn{1}{c}{\rotatebox[origin=l]{45}{\makebox[0pt][l]{\textbf{Collections}}}} & \multicolumn{1}{c}{\rotatebox[origin=l]{45}{\makebox[0pt][l]{\textbf{Citation Analysis}}}} & \multicolumn{1}{c}{\rotatebox[origin=l]{45}{\makebox[0pt][l]{\textbf{Trending Analysis}}}} & \multicolumn{1}{c}{\rotatebox[origin=l]{45}{\makebox[0pt][l]{\textbf{Author Profiles}}}} & \multicolumn{1}{c}{\rotatebox[origin=l]{45}{\makebox[0pt][l]{\textbf{Visualization Tools}}}} & \multicolumn{1}{c}{\rotatebox[origin=l]{45}{\makebox[0pt][l]{\textbf{Paper Chat}}}} & \multicolumn{1}{c}{\rotatebox[origin=l]{45}{\makebox[0pt][l]{\textbf{Idea Generation}}}} & \multicolumn{1}{c}{\rotatebox[origin=l]{45}{\makebox[0pt][l]{\textbf{Paper Writing}}}} & \multicolumn{1}{c}{\rotatebox[origin=l]{45}{\makebox[0pt][l]{\textbf{Summarization}}}} & \multicolumn{1}{c}{\rotatebox[origin=l]{45}{\makebox[0pt][l]{\textbf{Paper Review}}}} & \multicolumn{1}{c}{\rotatebox[origin=l]{45}{\makebox[0pt][l]{\textbf{Datasets}}}} & \multicolumn{1}{c}{\rotatebox[origin=l]{45}{\makebox[0pt][l]{\textbf{Code Repositories}}}} & \multicolumn{1}{c}{\rotatebox[origin=l]{45}{\makebox[0pt][l]{\textbf{LLM Integration}}}} & \multicolumn{1}{c}{\rotatebox[origin=l]{45}{\makebox[0pt][l]{\textbf{Web API}}}} & \multicolumn{1}{c}{\rotatebox[origin=l]{45}{\makebox[0pt][l]{\textbf{Personalization}}}} & \multicolumn{1}{c}{\textbf{Cost}} & \multicolumn{1}{c}{\textbf{Data Source}} \\ \midrule
\multirow{15}{*}{\centering \rotatebox[origin=c]{90}{\textbf{AI-Enhanced Search}}} & \href{https://elicit.com}{Elicit} & $\checkmark$ &  &  &  &  &  &  & $\checkmark$ & $\checkmark$ &  & $\checkmark$ & $\checkmark$ &  &  & $\checkmark$ &  &  & Freemium & 125 million  \\ 
& \href{https://openscholar.allen.ai}{OpenScholar} & $\checkmark$ &  & $\checkmark$ &  &  &  &  & $\checkmark$ &  &  & $\checkmark$ &  &  &  & $\checkmark$ &  &  & Free & 45 million \\ 
 & \href{https://www.undermind.ai/}{Undermind} & $\checkmark$ &  & $\checkmark$ &  &  &  &  & $\checkmark$ &  &  & $\checkmark$ &  &  &  & $\checkmark$ &  & $\checkmark$ & Premium & over 200 million \\ 
 & \href{https://www.perplexity.ai/}{Perplexity} & $\checkmark$ &  &  &  &  &  &  & $\checkmark$ & $\checkmark$ &  & $\checkmark$ & $\checkmark$ &  &  & $\checkmark$ &  &  & Freemium &  \\ 
 & \href{https://consensus.app}{Consensus} & $\checkmark$ &  & $\checkmark$ &  &  &  &  & $\checkmark$ &  &  & $\checkmark$ &  &  &  & $\checkmark$ & $\checkmark$ &  & Freemium & over 200 million \\ 
 & \href{https://typeset.io}{SciSpace} & $\checkmark$ &  & $\checkmark$ &  &  &  &  & $\checkmark$ & $\checkmark$ &  & $\checkmark$ & $\checkmark$ &  &  & $\checkmark$ &  &  & Freemium &  \\ 
 & \href{https://www.scienceos.ai/}{scienceQA} & $\checkmark$ &  & $\checkmark$ & $\checkmark$ &  &  &  & $\checkmark$ & $\checkmark$ &  & $\checkmark$ & $\checkmark$ &  &  & $\checkmark$ &  &  & Freemium & 220 million \\ 
 & \href{https://github.com/Future-House/paper-qa}{PaperQA2} &  &  &  &  &  &  &  & $\checkmark$ &  &  &  &  &  & $\checkmark$ & $\checkmark$ &  &  & Free &  \\ 
 & \href{https://paperguide.ai/}{Paperguide} & $\checkmark$ &  & $\checkmark$ &  &  &  &  & $\checkmark$ & $\checkmark$ &  & $\checkmark$ & $\checkmark$ &  &  & $\checkmark$ &  &  & Freemium &  \\ 
 & \href{https://www.hyperwriteai.com/}{HyperWrite} & $\checkmark$ &  &  &  &  &  &  & $\checkmark$ & $\checkmark$ & $\checkmark$ & $\checkmark$ & $\checkmark$ &  &  & $\checkmark$ &  &  & Premium &  \\ 
 & \href{https://www.researchkick.com/chat}{ResearchKick} & $\checkmark$ &  &  &  &  &  &  & $\checkmark$ & $\checkmark$ & $\checkmark$ & $\checkmark$ & $\checkmark$ &  &  & $\checkmark$ &  & $\checkmark$ & Premium &  \\ 
 & \href{https://www.bohrium.com}{Bohrium} & $\checkmark$ &  & $\checkmark$ &  &  & $\checkmark$ &  & $\checkmark$ &  &  &  &  &  &  & $\checkmark$ &  &  & Freemium & 170 million   \\
   & \href{https://paperpal.com}{Paperpal} & $\checkmark$ &  & $\checkmark$ &  &  &  &  & $\checkmark$ &  & $\checkmark$ & $\checkmark$ &  &  &  & $\checkmark$ &  &  & Freemium & over 3 million \\
    & \href{https://scholar.google.com/scholar_labs/search}{Scholar Labs} & $\checkmark$ &  &  &  &  &  &  &  &  &  &  &  &  &  & $\checkmark$ &  &  & Freemium &  \\
 \midrule
 \multirow{7}{*}{\centering \rotatebox[origin=c]{90}{\textbf{Graph-Based}}} & \href{https://www.connectedpapers.com/}{Connected Papers} & $\checkmark$ &  & $\checkmark$ &  &  &  & $\checkmark$ &  &  &  &  &  &  &  &  &  &  & Freemium & 214 million \\ 
 & \href{https://scholargps.com/}{ScholarGPS} & $\checkmark$ &  &  & $\checkmark$ & $\checkmark$ & $\checkmark$ & $\checkmark$ &  &  &  &  &  &  &  &  &  &  & Free & over 200 million \\ 
 & \href{https://citespace.podia.com/}{CiteSpace} &  &  &  &  & $\checkmark$ &  & $\checkmark$ &  &  &  &  &  &  &  &  &  &  & Freemium &  \\ 
 & \href{https://sci2.cns.iu.edu/user/index.php}{Sci2} &  &  &  &  &  &  & $\checkmark$ &  &  &  &  &  &  &  &  &  &  & Free &  \\ 
 & \href{https://nlpkg.sebis.cit.tum.de/}{NLP KG} & $\checkmark$ &  & $\checkmark$ & $\checkmark$ &  & $\checkmark$ & $\checkmark$ &  &  &  &  &  &  &  &  &  &  & Free &  \\ 
 & \href{https://ask.orkg.org}{ORKG ASK} & $\checkmark$ &  & $\checkmark$ &  &  &  &  &  &  &  & $\checkmark$ &  &  &  & $\checkmark$ &  &  & Free & 76 million \\ 
  & \href{https://www.litmaps.com}{Litmaps} & $\checkmark$ &  & $\checkmark$ &  &  &  & $\checkmark$ &  &  &  &  &  &  &  & $\checkmark$ &  &  & Freemium &  \\ 
  \midrule
 \multirow{11}{*}{\centering \rotatebox[origin=c]{90}{\textbf{Paper Chat}}} & \href{https://chatgpt.com/}{ChatGPT} & $\checkmark$ &  &  &  &  &  &  & $\checkmark$ & $\checkmark$ & $\checkmark$ & $\checkmark$ & $\checkmark$ &  &  & $\checkmark$ & $\checkmark$ &  & Freemium & 10 PDF files \\ 
 & \href{https://claude.ai/}{Claude} & $\checkmark$ &  &  &  &  &  &  & $\checkmark$ & $\checkmark$ & $\checkmark$ & $\checkmark$ & $\checkmark$ &  &  & $\checkmark$ & $\checkmark$ &  & Freemium & 5 PDF files \\ 
 & \href{https://chat.deepseek.com}{Deepseek} & $\checkmark$ &  &  &  &  &  &  & $\checkmark$ & $\checkmark$ & $\checkmark$ & $\checkmark$ & $\checkmark$ &  &  & $\checkmark$ & $\checkmark$ &  & Free &  \\ 
 & \href{https://un.ms/research}{Research} &  &  & $\checkmark$ &  &  &  &  & $\checkmark$ & $\checkmark$ &  & $\checkmark$ & $\checkmark$ &  &  & $\checkmark$ &  &  & Freemium & 1 PDF file \\ 
 & \href{https://notebooklm.google}{NotebookLM} &  &  &  &  &  &  &  & $\checkmark$ & $\checkmark$ &  & $\checkmark$ & $\checkmark$ &  &  & $\checkmark$ &  & $\checkmark$ & Freemium & 50 PDF files \\ 
 & \href{https://www.read.enago.com}{Enago Read} & $\checkmark$ &  & $\checkmark$ &  &  &  &  & $\checkmark$ & $\checkmark$ &  & $\checkmark$ & $\checkmark$ &  &  & $\checkmark$ &  & $\checkmark$ & Freemium & 1 PDF file \\ 
 & \href{https://docanalyzer.ai/}{DocAnalyzer.AI} &  &  & $\checkmark$ &  &  &  &  & $\checkmark$ & $\checkmark$ &  & $\checkmark$ & $\checkmark$ &  &  & $\checkmark$ & $\checkmark$ & $\checkmark$ & Premium & few PDF files \\ 
 & \href{https://www.getcoralai.com/}{CoralAI} &  &  & $\checkmark$ &  &  &  &  & $\checkmark$ & $\checkmark$ &  & $\checkmark$ & $\checkmark$ &  &  & $\checkmark$ &  &  & Freemium & 1 PDF file \\ 
 & \href{https://www.explainpaper.com/}{ExplainPaper} &  &  &  &  &  &  &  & $\checkmark$ & $\checkmark$ &  & $\checkmark$ & $\checkmark$ &  &  & $\checkmark$ &  &  & Freemium & 1 PDF file \\ 
 & \href{https://www.chatPDF.com/}{ChatPDF} & $\checkmark$ &  & $\checkmark$ &  &  &  &  & $\checkmark$ & $\checkmark$ &  & $\checkmark$ & $\checkmark$ &  &  & $\checkmark$ &  &  & Premium & 1 PDF file \\ 
  & \href{https://answerthis.io}{AnswerThis} & $\checkmark$ &  & $\checkmark$ &  &  &  &  & $\checkmark$ &  & $\checkmark$ & $\checkmark$ & $\checkmark$  &  &  & $\checkmark$ &  &  & Freemium & over 300 million \\
 \midrule
 \multirow{7}{*}{\centering \rotatebox[origin=c]{90}{\textbf{Recommender}}} & \href{https://arxiv-sanity-lite.com/}{Arxiv Sanity} & $\checkmark$ & $\checkmark$ & $\checkmark$ &  &  &  &  &  &  &  &  &  &  &  &  &  & $\checkmark$ & Free &  \\ 
 & \href{https://www.scholar-inbox.com/}{Scholar Inbox} & $\checkmark$ & $\checkmark$ & $\checkmark$ &  & $\checkmark$ &  & $\checkmark$ &  &  &  &  &  &  &  & $\checkmark$ &  & $\checkmark$ & Free &  \\ 
 & \href{https://researchtrend.ai}{ResearchTrend.ai} & $\checkmark$ &  &  &  & $\checkmark$ &  &  &  &  &  &  &  &  &  &  &  &  & Freemium &  \\ 
 & \href{https://trendingpapers.com/}{TrendingPapers} & $\checkmark$ & $\checkmark$ &  &  & $\checkmark$ &  &  &  &  &  & $\checkmark$ &  &  &  & $\checkmark$ &  & $\checkmark$ & Free &  \\ 
 & \href{https://dev.bytez.com/}{Bytez} & $\checkmark$ &  &  &  & $\checkmark$ &  &  & $\checkmark$ & $\checkmark$ &  & $\checkmark$ & $\checkmark$ &  &  & $\checkmark$ & $\checkmark$ &  & Freemium &  \\ 
 & \href{https://notesum.ai/}{Notesum.ai} & $\checkmark$ & $\checkmark$ & $\checkmark$ &  &  &  &  &  &  &  & $\checkmark$ &  &  &  & $\checkmark$ &  & $\checkmark$ & Freemium &  \\ 
 & \href{https://www.researchrabbit.ai/}{Research Rabbit} & $\checkmark$ &  & $\checkmark$ &  &  &  & $\checkmark$ &  &  &  &  &  &  &  &  &  &  & Free &  \\
\end{tabular}%
}
\end{changedfloat}
\end{table}


%% file: topics/experiments.tex
\subsection{AI-Driven Scientific Discovery: Ideation, Hypothesis Generation, and Experimentation}
\label{sec:experiments}

\mybox{Ideation
focuses on proposing new tools and\slash or analyzing existing ones, while hypothesis generation involves formulating specific, testable questions that guide empirical or theoretical justifications.  
In today's age of rapidly growing scientific literature, the effort of moving from literature review to idea or hypothesis formation has become increasingly time-consuming. 
Recently, LLMs have been employed to address this issue by making idea and hypothesis formation efficient: \changed{they are being leveraged both as generators (to autonomously produce ideas and hypotheses) and as evaluators (to assess their quality and select those that are meaningful, relevant, and novel).}
Experimentation adds further complexity, requiring
careful methodological design, large-scale simulations, and in-depth results analysis. \changed{In this section, we first review ideation, hypothesis generation, and their (intrinsic and downstream) evaluation, then discuss how experimentation, framed as a form of downstream evaluation, can be automated through LLMs.}}

\subsubsection{Data}

\begin{table}
\begin{changedfloat}
\caption{Overview of datasets for idea and hypothesis generation and experimentation}
\label{tab:section4.2_dataset}
\centering 
\small
\setlength\tabcolsep{4pt} 
\begin{tabular}{lccccc}
\toprule
\textbf{Dataset} & \textbf{Source} & \textbf{Data Size} & \textbf{Domain} & \textbf{Time Span} & \textbf{Task}\\ 
\midrule
SciMON~\cite{chai2024exploring} & ACL Anthology & 135,814 papers &  NLP   & 1952--2022 & Idea Generation\\ 
IDEA Challenge~\cite{ege2023idea} & University of Bristol & 240 prototypes &  Engineering  & 2022 & Idea Generation\\ 
SPACE-IDEAS+~\cite{garcia-silva-etal-2024-space} & COLING & 1020 ideas &  Physics  & 2024 & Idea Generation\\ 
TOMATO-Chem~\cite{yang2024moose} & Nature and Science & 51 papers & Chemistry & 2024 & Hypothesis Generation\\
LLM4BioHypoGen~\cite{qi2024large} & PubMed & 2,900 papers & Medicine & 2000--2024 & Hypothesis Generation\\
CSKG-600~\cite{dessi2022cs} & CSKG & 600 hypotheses & AI & 2010--2017& Hypothesis Generation \\
ScienceAgentBench~\cite{chen2024scienceagentbenchrigorousassessmentlanguage} & OSU NLP & 44 papers & Diverse
 & 2024 & Automated Experimentation\\
SWE-bench~\cite{jimenez2024swebench} & ICLR & 2,294 issues & SWE & 2024 & Automated Experimentation\\
MLGym-Bench~\cite{nathani2025mlgymnewframeworkbenchmark} & Meta & 13 tasks & Diverse
& 2025 & Automated Experimentation\\
\bottomrule
\end{tabular}
\end{changedfloat}
\end{table}

Here we survey diverse datasets for evaluating LLMs in hypothesis generation, idea formation, and experimentation.
These datasets, summarized in Table~\ref{tab:section4.2_dataset}, were constructed from various scholarly sources representing a variety of scientific domains.

\begin{changedpars}
    \textbf{SciMON}~\cite{chai2024exploring}, a dataset for the idea generation task, is a subset of the Semantic Scholar Open Research Corpus (S2ORC)~\cite{lo2019s2orc} focusing on abstracts of Association for Computational Linguistics~(ACL) publications from 1952 to 2022. It contains 135,814 abstracts, divided into training (before 2021), validation (2021), and test (2022) sets. Each abstract was annotated using PL-Marker~\cite{ye2021packed} and a structure classifier~\cite{cohan2019pretrained} that extract keywords and categorize sentences as providing background, research ideas, etc. The tools were evaluated on a human-curated subset, and only high-confidence annotations were retained; \changed{despite this, some annotations may be incorrect, and errors can propagate and affect ideation quality.}

    \textbf{SPACE-IDEAS+}~\cite{garcia-silva-etal-2024-space} releases two versions of datasets. The smaller one contains 176 ideas sampled from the Open Space Innovation Platform (OSIP), an online repository of publicly available ideas related to space innovation. All the ideas were manually annotated by human experts, where each sentence was labeled with one of five roles (Challenge, Proposal, Elaboration, Benefits, Context) by two human annotators. Meetings were conducted to resolve disagreements between annotators. The larger version contains 1,020 ideas that were annotated by having GPT-3.5-turbo adopt the same annotation guidelines. A subset of the generated annotations was evaluated by comparing with human annotations; the agreement of 50\% indicates the mediocre quality of GPT annotations. 

    \textbf{TOMATO-Chem}~\cite{yang2024moose} is a hypothesis generation dataset containing 51 chemistry and material science papers published in \emph{Nature} or \emph{Science} in 2024. To these papers experts applied annotations concerning the background, research questions, works that potentially inspired the paper, hypotheses, and experiments for hypothesis justification. \changed{Details concerning the annotation task (e.g., the number of annotators and their agreement) are unfortunately not reported.}

    \textbf{LLM4BioHypoGen}~\cite{qi2024large}, another hypothesis generation dataset, consists of 2,900 medical publications sourced from PubMed, where 2,500 papers were used for the training set and 200 papers for validation set (both published before January 2023), with 200 papers in the test set (published after August 2023). 
    Each paper was annotated by using GPT to construct background and hypothesis pairs; however, no human evaluation of these pairs was provided.

\textbf{ScienceAgentBench}~\cite{chen2024scienceagentbenchrigorousassessmentlanguage} is an automated experimentation dataset
comprised of 102 tasks derived from 44 peer-reviewed publications across four disciplines: bioinformatics, computational chemistry, geographical information science, and psychology \& cognitive neuroscience. Each task requires an agent to generate a self-contained Python program based on a natural language instruction, a dataset, and optional expert-provided knowledge. The benchmark employs multiple evaluation metrics, including the valid execution rate, success rate, CodeBERTScore, and API cost, to assess the generated programs' correctness, execution, and efficiency.

    \textbf{SWE-bench}~\cite{jimenez2024swebench} is a similar dataset 
    containing 2,294 tasks derived from 12 popular open-source Python repositories in real-world software engineering.  
    Each task requires the model to edit a full codebase based on an issue description, producing a patch that must apply cleanly and pass fail-to-pass tests. The benchmark features long, complex inputs, robust evaluation via real-world testing, and the ability to be continually updated with new issues. However the execution-based testing can be misleading as it does not assess criteria such as comprehensiveness, efficiency, or readability. While there is an additional rubric-based human evaluation where final versions are revised by experts, these human annotators are mainly familiar with Python and tend to dismiss other languages.
\end{changedpars}

\subsubsection{Methods and Results}
\begin{figure*}
  \begin{changedfloat}
  \centering
  \includegraphics[width=\textwidth]{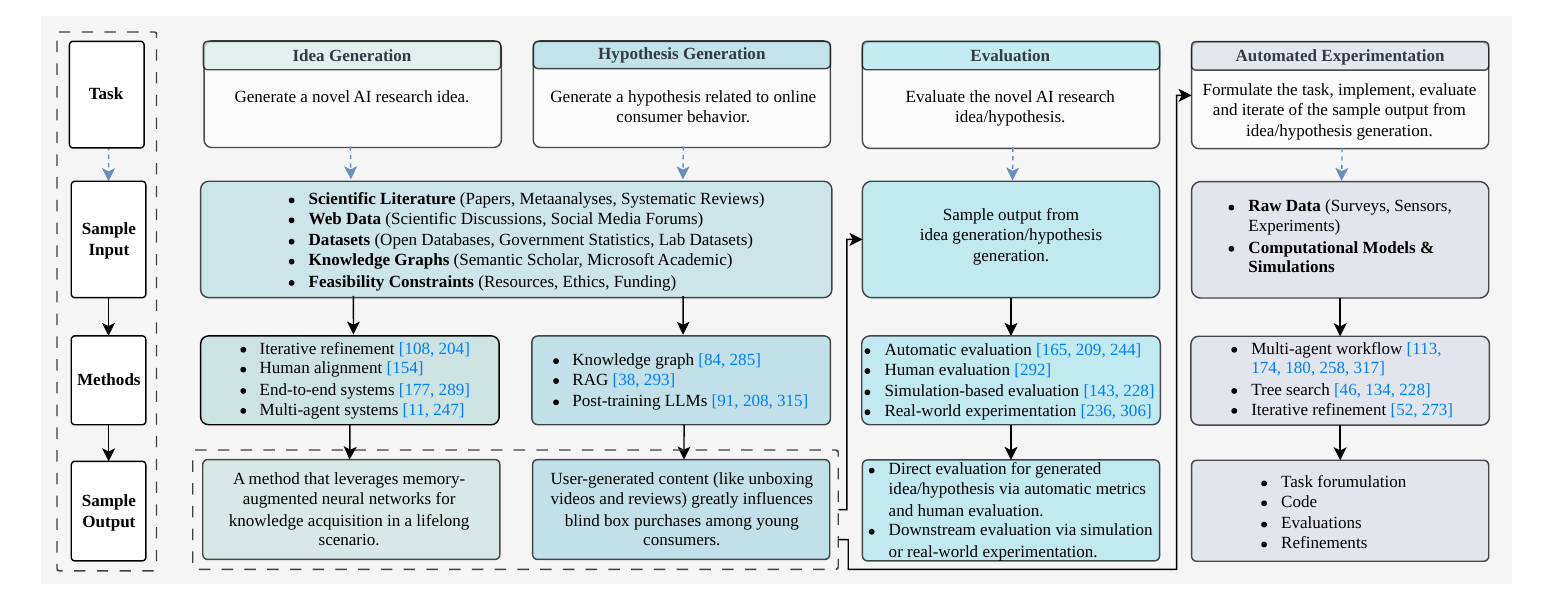} 
  \caption{Examples of idea generation, hypothesis generation, evaluation, and automated experimentation follow a four-component structure: \emph{task}, \emph{sample input}, \emph{methods}, and \emph{sample output}. The \emph{task} defines the goal of each process. For idea\slash hypothesis generation, \emph{sample input} consists of benchmark datasets for each task, whereas evaluation uses its \emph{sample output}.
  \changed{Automated experimentation uses \emph{sample output} from idea\slash hypothesis generation for the \emph{task} and raw data\slash computational models and simulation as \emph{sample input}. \emph{Methods} encompass a taxonomy of approaches. \emph{Sample output} differs by process: idea\slash hypothesis generation and automated experimentation yield textual outputs (descriptions, explanations, or code), whereas evaluation produces quantitative scores.}}
  \label{fig:scientific_discovery}
  \end{changedfloat}
\end{figure*}

Here, we discuss state-of-the-art methods and results in hypotheses generation, idea formation,
and automated experimentation. 
Figure~\ref{fig:scientific_discovery} provides some examples for each approach.

\paragraph{Idea Generation} 
Several methods have been proposed to generate research ideas,
variously using iterative refinement, 
human alignment, \changed{end-to-end autonomous systems,} or multi-agent systems~\cite{lu2024aiscientist, su2024two, radensky2024scideator, hu2024nova, li2024chain, baek2025researchagent, weng2025cycleresearcher, yamada2025ai, jansen2025codescientist,kumbhar2025hypothesis,schmidgall2025agent}. 
    For \textbf{iterative refinement},~\citet{hu2024nova} introduce an iterative planning and search framework aimed at enhancing the novelty and diversity of ideas generated by LLMs. By systematically retrieving external knowledge, the approach addresses the limitations of existing models in producing simplistic or repetitive suggestions. 
    Similarly, IdeaSynth~\cite{pu2024ideasynth} focus on iterative refinement by providing literature-grounded feedback. It represents research ideas as nodes on a canvas to facilitate the idea iteration and composition of different idea facets, enabling researchers to develop more detailed and diverse ideas.
    Work involving \textbf{human alignment}
    seeks to organize information in ways that mirror human research processes. Chain of Ideas (CoI)~\cite{li2024chain} proposes structuring literature into a chain to emulate the progressive development of research domains. This facilitates the identification of meaningful directions and has been shown to outperform existing methods in generating ideas comparable in quality to those produced by human researchers. Scideator~\cite{radensky2024scideator}, in contrast, focuses on recombining facets (e.g., purposes, mechanisms, and evaluations) from existing research papers to synthesize novel ideas. By incorporating automated novelty assessments, Scideator enables users to identify overlaps and refine their ideas. 
    \changed{The AI Scientist~\cite{lu2024aiscientist} is a prominent example of an \textbf{end-to-end autonomous system}. It is a framework for automating the entire research pipeline, including idea generation, experiment execution, and paper writing. The AI Scientist-v2~\cite{yamada2025ai} employs an agentic framework using tree search and iterative refinement.}
    Among \textbf{multi-agent systems},
    VirSci~\cite{su2024two} employs an ensemble 
    of virtual agents to collectively generate, evaluate, and refine ideas. 
    It outperforms individual LLMs, underscoring the potential of teamwork in enhancing scientific innovation.
ResearchAgent~\cite{baek2025researchagent} is another a multi-agent LLM framework that enables ideation by generating new research problems, methods, and experiments from existing literature and iteratively refining them through LLM-assisted peer review. 

\paragraph{Hypothesis Generation}

A considerable number of studies~\cite{yang2023large, zhou2024hypothesis, tong2024automating, bai2024advancing, wang2024llm, park2024can, qi2023large, chen2024use, liu2024beyond, banker2024machine, tang2023less, yang2024moose, chen2023chemist, jing2024data,xiong2024improving,ghafarollahi2025sciagents, zhou2024hypothesis, gottweis2025towards, movva2025sparse, liu2025hypobench, abdel2025scientific} have employed LLMs to generate hypotheses, all of which differ in their approaches.  \changed{These can be broadly categorized as employing (a)~knowledge graphs, (b)~retrieval-augmented generation, and\slash or (c)~post-training of LLMs.
    Among \textbf{knowledge graph} approaches is KG-CoI~\cite{xiong2024improving}, which uses an external KG that grounds the model's reasoning in structured knowledge, enhancing the quality of generated hypotheses and reducing hallucination during hypothesis generation. SciAgents~\cite{ghafarollahi2025sciagents} is a framework that combines KGs with multi-agent systems and LLMs; it organizes scientific concepts into a graph and enables agents to reason over this structure to generate and iteratively refine hypotheses.
    For \textbf{RAG}, MOOSE-Chem~\cite{yang2024moose} is a retrieval-based system for the rediscovery of high-impact chemistry hypotheses given only a research background. It first retrieves relevant papers from a large corpus, then uses these papers together with the background to generate candidate hypotheses, and finally ranks the hypotheses by quality. Chemist-X~\cite{chen2023chemist} is retrieval-augmented agent for reaction condition discovery in chemical synthesis. It uses LLMs to retrieve information (e.g., reaction conditions and molecules) from chemical literature and molecular databases, helping to narrow the search space for plausible reaction conditions, which are then treated as candidate hypotheses about optimal experimental settings.
    Major strategies for \textbf{post-training LLMs},
    include few-shot learning, fine-tuning, and iterative refinement. \citet{qi2023large} 
    find that hypotheses generated by few-shot learning
    are judged by humans as more testable than those generated in a zero-shot setup. They report that while fine-tuning improves the overall quality of hypotheses, the improvement is limited to the domain of training data; in unseen domains, fine-tuning harms hypothesis quality, particularly the novelty aspect. \citet{zhou2024hypothesis} iteratively refine hypotheses through reinforcement learning, aiming to
    increase the similarity between a research problem and a generated hypothesis. The AI co-scientist~\cite{gottweis2025towards} employs a multi-agent system relying on a generate--debate--evolve framework to iteratively enhance generated hypotheses.}

\paragraph{Evaluation} The evaluation of generated ideas and hypotheses ensures that they are scientifically meaningful and feasible. \changed{Evaluation approaches \cite[e.g.,][]{yang2024moose, chai2024exploring, starnes2024mamba, qi2024large, yang2023large, laurent2024lab, schmidgall2024agentclinic, zhang2025integrating, si2025ideation, wang2024scimon} can be generally categorized as intrinsic (automatic or human evaluation) or downstream (simulation-based evaluation or real-world experimentation).}

    \changed{For \textbf{automatic evaluation}, metrics such as ROUGE~\cite{chin2004rouge} have been used to assess the quality of generated hypotheses (or ideas) by measuring their similarity to human-annotated ground truths~\cite{starnes2024mamba}. \citet{qi2024large} leverage LLMs-as-judge to evaluate hypotheses based on four scientific aspects: novelty, relevance to the given background, significance within the research community, and verifiability (i.e., testability). 
    In \textbf{human evaluation}, domain experts are involved to assess the quality of generated ideas and hypotheses, especially when ground truth is unavailable. For instance, \citet{yang2023large} invited social scientists to provide feedback on the LLM-generated hypothesis, ``In collectivist cultures, individuals engage in more conspicuous consumption behaviors compared to individualistic cultures.'' The social scientists found the hypothesis potentially novel and counterintuitive, as prior research suggests that collectivist cultures typically discourage conspicuous displays of personal wealth. Such expert feedback helps identify hypotheses that are meaningful and worth further investigation.
    LabBench~\cite{laurent2024lab} and AgentClinic~\cite{schmidgall2024agentclinic} are examples of \textbf{simulation-based evaluation}, where
    generated hypotheses are evaluated in virtual laboratory environments. These tools simulate laboratory conditions in materials science and biomedicine, allowing hypotheses to be tested cost-effectively. 
    \textbf{Real-world experimentation} is employed by \citet{zhang2025integrating}, who evaluate candidate hypotheses (protein variants) within an automated biofoundry. Each variant is constructed and evaluated using instruments such as liquid handlers and thermocyclers, 
    together with peripheral devices like plate sealers, shakers, and incubators, to measure enzymatic activity, protein yield, and success rates. \citet{si2025ideation} invited 43 expert researchers to experiment with LLM-generated ideas. Each researcher spent over 100 hours implementing the assigned ideas and wrote a short paper to report the results. Ideas proposed by human researchers were also implemented as a control group. All the papers were then reviewed blindly by expert reviewers. Experimental results from LLM-generated ideas were judged by expert reviewers to be less novel, exciting, or effective compared to those from human ideas.}

\paragraph{Automated Experimentation.} 
Experimentation, which can serve as downstream evaluation for empirically validating AI-generated ideas and hypotheses, encompasses task formulation, implementation, evaluation, and iteration. Automated experimentation aims to streamline this workflow, with approaches like neural architecture search~\cite{elsken2019neural} and AutoML~\cite{he2021automl}. LLMs further enhance this by enabling automation through natural language prompts. AutoML-GPT~\cite{tsai2023automl} and MLcopilot~\cite{zhang2023mlcopilot} use LLMs for hyperparameter tuning, while MLAgentBench~\cite{huang2024mlagentbench} benchmarks fundamental automation tasks. Recent work explores advanced frameworks incorporating multi-agent collaboration, tree search, and iterative refinement for scientific experimentation.

    For \textbf{multi-agent workflow}, GVIM~\cite{ma2025gvim} enhances chemical research with domain-specific functions, while DrugAgent~\cite{liu2024drugagent} employs LLMs for task planning in drug discovery. AutoML-Agent~\cite{trirat2024automl} integrates retrieval-augmented planning for AutoML tasks, and MLAgentBench~\cite{huang2024mlagentbench} benchmarks LLM-driven agents in machine learning experimentation. The Agent-as-a-Judge framework~\cite{zhuge2024agent} introduces structured agent evaluation. 
    For \textbf{tree search}, AIDE~\cite{schmidt2024introducing} applies solution space tree search to refine solutions in Kaggle challenges. The Tree Search for Language Model Agents framework~\cite{koh2024tree} enables LLM agents to plan multi-step interactions using best-first tree search, pruning less promising options. SELA~\cite{chi2024sela} combines LLM-generated insights with Monte Carlo tree search, iteratively refining machine learning experiments by selecting promising configurations and executing them.
    For \textbf{iterative refinement}, APEx~\cite{conti2024benchmarking} automates LLM-based experimentation with an orchestrator, execution engine, benchmark generator, and model library. OpenHands~\cite{wang2024openhands} enables AI agents to interact with software, execute actions in a sandboxed runtime, and collaborate across tasks using predefined benchmarks.

\subsubsection{Domains of Application}

Studies have addressed ideation and hypothesis generation in NLP, engineering, physics, chemistry, the social sciences, and medicine.  
\changed{Work on automated experimentation has similarly relied on domain-specific datasets to guide the process of designing and testing experiments.
However, the designs underlying these systems are typically domain-agnostic. For instance, iterative refinement, human alignment, multi-agent systems, and tree search are low-level methodologies that are applicable across multiple domains. Regarding evaluation, except for automatic evaluation, most evaluation approaches are limited to the domains for which they are designed: human evaluation relies on domain-specific evaluation guidelines, simulation-based evaluation requires domain-specific laboratory configurations, and infrastructure in which real-world experimentation is conducted may differ across domains.}

\subsubsection{Limitations and Future Directions}

A large-scale study~\cite{si2024llmsgeneratenovelresearch} comparing human researchers and LLMs finds that LLMs generate ideas judged to be more novel but slightly less feasible, highlighting challenges like limited diversity and self-evaluation failures.
Additionally, 
given that ideas and hypotheses are theoretical and costly to validate, it is unclear whether they could lead to scientific discovery.
Furthermore, previous methods lack due diligence through data, 
and therefore generated ideas and hypotheses are often too general~\cite{yang2024moose}. Moreover, LLMs may end up re-generating recently discovered ideas and hypotheses, since they lack access to recent scientific papers~\cite{liu2024beyond}. Their outputs are moreover very sensitive to the framing of input prompts~\cite{park2024can}. Future work could focus on improving feasibility and diversity of ideas and hypotheses, consulting scientific papers in real time, and refining ideation and hypothesis generation through data inspection. 
LLM-automated experimentation has several additional challenges. First, LLMs' propensity to hallucinate results or references disrupt the precise steps required for experimental workflows. They also struggle to integrate and align different modalities, such as video, audio, or sensory data, which are increasingly essential in modern scientific experimentation. Moreover, LLMs lack the critical analysis capabilities necessary to identify flaws or refine hypotheses during experimentation. Particularly in biology and chemistry, they may also struggle with precise reasoning and tool usage, which are vital for ensuring experimental success~\cite{reddy2024scientificdiscoverygenerativeai}.


%% file: topics/textgeneration.tex
\subsection{Text-based Content Generation}
\label{sec:textgeneration}

\mybox{\changed{Generating scientific content includes generating texts of various types and lengths, each demanding different skills. In this section, we focus on selected subtasks to emphasize the varied challenges inherent in each. \textbf{Titles} of scientific papers need to reflect the content of a paper in a few catchy words, while \textbf{abstracts} are concise, stand-alone summaries. Approaches to generating \textbf{long texts} face challenges such as structuring arguments and maintaining factual consistency. \textbf{Related work} generation also requires text summarization skills, though in a more concise form. 
Generating \textbf{bibliographic references} depends on scientific discovery and literature research and has limited room for variation in phrasing, unlike in tasks such as \textbf{proof-reading and paraphrasing}. 
As we discuss, current generative models exhibit varying performance across these subtasks.}}

\subsubsection{Data}

Open access research articles are a valuable data source for text-based content generation. These include scientific publisher repositories offering at least some open access content (e.g., \href{https://www.nature.com/nature-portfolio/for-authors/nature-research-journals}{Nature portfolio}, \href{https://www.tandfonline.com/}{Taylor \& Francis}) as well as preprint repositories (e.g., \href{https://arxiv.org/}{arXiv}, \href{https://www.biorxiv.org/}{bioRxiv}).
These repositories can be leveraged to develop datasets with pairs of titles and abstracts or abstract and conclusion/future work pairs. \citet{wang-etal-2019-paperrobot}, for example, extract from PumMed title to abstract pairs, abstract to conclusion and future work pairs, and conclusion and future work to title pairs. Annotated, task-specific datasets for scientific text generation 
are presented in
Table~\ref{tab:data_text_generation}.

\begin{table}
\begin{changedfloat}
\small
    \caption{Annotated or task-specific datasets for scientific text generation.}
    \label{tab:data_text_generation}
    \centering
    \begin{tabular}{lp{4.8cm}p{2.6cm}p{3.2cm}}
    \toprule
       \textbf{Dataset}  & \textbf{Size} & \textbf{Sources} & \textbf{Application} \\
       \midrule
	   Abstract-title humor~\cite{chen-eger-2023-transformers} & 2,638 humor annotated titles & ML and NLP domain & Title generation\\
	   PaperRobot~\cite{wang-etal-2019-paperrobot} & 27K title--abstract, 27K abstract--conclusion\slash future work, 20K conclusion\slash future work--title pairs & PubMed & Title, abstract, conclusion, and future work generation\\
        ScisummNet~\cite{yasunaga2019scisummnet} & 1,000 papers + 20 citation sentences each & ACL Anthology & Related work generation\\
CORWA~\cite{li-etal-2022-corwa} & 927 related work sections & NLP domain & Related work generation\\
		CiteBench~\cite{funkquist-etal-2023-citebench} & 358,765 documents + citations & arXiv et al. & Related work generation\\
        LongWriter~\cite{bai2024longwriter} & 6,000 texts (literature, academic writing, popular science, news) & Open datasets & Long text generation\\
        LongWriter-Zero~\cite{wu2025longwriterzero} & 8,610 instruction tuning requiring outputs exceeding 10,000 words & Open datasets & Long text generation\\
        LongEval~\cite{wu2025longeval} & 166 high-quality human-authored samples exceeding 2,000 words & arXiv, blogs, and Wikipedia & Long text generation  \\
        Casimir~\cite{jourdan2024casimir} & 15,646 papers (consecutive versions) & OpenReview & Paraphrasing \\
        ParaRev~\cite{jourdan2025pararev} & 48,203 paper paragraphs (consec. versions) & OpenReview & Paraphrasing \\
\bottomrule
    \end{tabular}
\end{changedfloat}
\end{table}

\subsubsection{Methods and Results}

Here we survey approaches to generating 
textual content for scientific papers, 
such as titles, abstracts, related work sections, and bibliographies. 
An overview of these processes is given in Appendix~\ref{ax:content_generation}. 

\paragraph{Title Generation.} 
Generating appropriate titles for scientific papers is an important task because titles are the first access point of a paper and can attract substantial reader interest; titles can also influence the reception of a paper~\cite{letchford2015advantage}.  Consequently, several works have targeted generating titles automatically, often using paper abstracts as input. For example, \citet{mishra2021automatic} use a pipeline of three modules, viz.\ generation by transformer based (GPT-2) models, selection (from multiple candidates) and refinement. 
\citet{chen-eger-2023-transformers} also leverage transformers for title generation from abstracts, including humorous title generation. Their results show that none of the applied models (BART, GPT-2, T5, GPT-3.5) can adequately generate humorous titles.
\changed{Sebo et al.~\cite{sebo2025can} find that GPT-4o generated titles are preferred by human raters over human titles, given the abstract.}
\citet{wang-etal-2019-paperrobot} address the problem differently by drafting titles based on future work sections of previous related papers.

\paragraph{Abstract Generation}
There are several approaches trying to assess the capabilities of LLMs to generate abstracts based on context information such as paper titles, journal names, keywords, or the full text of the paper. \citet{hwang2024can} assess the ability of GPT-3.5 and -4 to generate abstracts based on a full text. The results are manually evaluated using the Consolidated Standards of Reporting Trials for abstracts, a standard that aims to enhance the overall quality of scientific abstracts~\cite{hopewell2008consort}. 
While the readability of GPT-generated abstracts is rated higher, their overall quality is inferior to the originals. 
\citet{wang-etal-2019-paperrobot} generate abstracts from titles, leveraging transformers and knowledge bases. \citet{gao2023comparing} collect 50 publications from five medical journals and use ChatGPT to generate abstracts based on their titles and journal names. Both original and generated abstracts are evaluated using AI output detectors and human reviewers. Human reviewers are able to identify 68\% of the generated abstracts, but misclassified 14\% of original abstracts as LLM-generated.  
\citet{farhat2023trustworthy} evaluate the performance of ChatGPT generating abstracts based on three keywords, the name of a database (Scopus or Web of Science), and the task to analyze bibliographic data in the domain indicated by the keywords. 
\changed{Manually comparing the generated abstracts to original abstracts on the same topic, the authors conclude that at the time of the study, 
ChatGPT was not a trustworthy tool.}

\paragraph{Long Text Generation}  

The AI Scientist~\cite{lu2024aiscientist,yamada2025ai} produces complete scientific manuscripts by leveraging outputs from earlier stages of the scientific lifecycle, such as experimental results, intermediate analyses, and candidate citations. By conditioning
on these intermediate outputs, the system drafts papers largely conforming to domain-specific conventions, including citation and disciplinary writing norms. 
\changed{However, 
the system does not explicitly address the challenge of maintaining global narrative coherence, nor does it provide principled mechanisms for modeling long-range logical dependencies and argument consistency across extended texts.}
Beyond end-to-end research automation frameworks, other work focuses more specifically on long-form text generation itself. 
LongWriter~\cite{bai2024longwriter} addresses   
long-form text generation by targeting long-range coherence and structural consistency. 
The model introduces hierarchical attention mechanisms to enhance thematic consistency across extended texts and employs tailored fine-tuning strategies to better align generated outputs with user prompts. 
LongEval~\cite{wu2025longeval} provides a systematic evaluation of long-text generation under both direct and plan-based generation paradigms across academic, encyclopedic, and blog-style domains. 
The findings suggest that larger, general-purpose, instruction-tuned models often perform comparably to specialized, smaller long-text models (e.g., LongWriter), raising questions about the marginal benefits of domain-specific fine-tuning for long-form generation. 
\changed{Motivated by the need for stronger structural control, recent work has increasingly moved beyond standard supervised fine-tuning (SFT) toward approaches based on reinforcement learning (RL). 
LongWriter-Zero~\cite{wu2025longwriterzero} demonstrates that RL without SFT can enable ultra-long text generation (i.e., 10,000+ words). 
By employing composite reward models that jointly evaluate length, quality, and formatting constraints, LongWriter-Zero achieves competitive or even superior performance relative to proprietary models (i.e., Claude-Sonnet-4, Gemini-2.5-Pro) in long-form generation tasks.}
Similarly, LongReward~\cite{zhang2024longrewardimprovinglongcontextlarge} leverages RL with custom-designed reward signals that emphasize coherence, factual accuracy, and linguistic quality. These reward mechanisms are particularly relevant for scientific text generation, where accuracy and adherence to domain-specific conventions are crucial.

\paragraph{Related Work Generation}

There has been a substantial body of work on related work generation through text summarization, with variances in the approach (extractive or abstractive) and the citation text length (sentence- or paragraph-level). Extractive approaches focus on selecting sentences from cited papers and reordering them
to form a paragraph of related work. For instance, \citet{hoang-kan-2010-towards} propose an extractive summarization approach that selects sentences describing the cited papers. 
Subsequent extractive approaches differ from this approach in how they order the extracted sentences: while 
\citet{wang2019toc} assume that the sentence order is given, \citet{hu2014automatic} and \citet{deng2021automatic} take advantage of an automatic approach to reorder sentences based on topic coherence. 
Most abstractive approaches are based on language models and focus on generating either a single sentence from a single reference, 
or a single paragraph from multiple references. Typically, the abstractive process is repeated until a related work section is complete.
\citet{abura2020automatic} introduce an abstractive summarization approach to generate citation sentences in a single-reference setup. Their approach has been trained on the \textbf{ScisummNet} corpus~\cite{yasunaga2019scisummnet} with paper abstracts as inputs and citation sentences as outputs. Li et al.~\cite{li-etal-2022-corwa} further extend this idea to a multiple-reference setup, namely generating a paragraph of citation sentences from various cited papers. Their approach has been trained on the \textbf{CORWA} corpus~\cite{li-etal-2022-corwa} to generate both citation and transition sentences. 
Recently, several works have explored LLMs for related work generation. For instance, \citet{sahinuc-etal-2024-systematic} 
use 
instruction promoting with
LLMs, 
an alternative to extractive and abstractive approaches, to generate citation sentences. \citet{martin2024shallow} employ a citation graph coupled with LLMs to produce a related work section. \changed{\citet{li2025explaining} use LLM prompting to extract features that capture relationships between citing and cited papers; these features are then composed into a new prompt that enables the LLM to generate a related work section. \citet{csahinucc2025expert} introduce a multi-turn evaluation framework for assessing the quality of AI-generated related work sections. The framework uses expert preferences to align with human judgment, and iteratively evaluates section drafts and generates natural-language feedback for revision.}
Overall, 
extractive approaches, while factual, often lack fluency and coherence. In contrast, abstractive approaches and instruction prompting, which are based on (large) language models, do not struggle with these issues, however, they suffer from factual errors, known as hallucination.

\paragraph{Citation Generation}

Bibliographic references are important for ensuring the scientific integrity of a paper. However, in many cases, cited references generated by LLMs such as ChatGPT are reported not to exist---that is, they are hallucinated or incorrect~\cite{li-ouyang-2024-related,huang2023citation,
farhat2023trustworthy}. Most studies on this phenomenon are case studies presenting one or more examples. 
In a study by \citet{walters2023fabrication}, GPT-3.5 and -4 are used to generate 84 literature reviews containing 636 bibliographic citations. 
55\% of the GPT-3.5 citations were fabricated, compared to 18\% of GPT-4's. Additionally, 43\% of non-fabricated GPT-3.5 citations contain substantive citation errors, compared to 24\% for GPT-4. Despite this notable improvement, issues with citation fabrication and errors persist. 
\changed{In ScholarPilot~\cite{wang2025scholarcopilot}, retrieval tokens are generated to query citation databases and the retrieved references are directly fed into the model to augment the text generation process. The model is based on Qwen-2.5-7B and fine-tuned on a corpus of 500K arXiv papers. It outperforms other Qwen-2.5 variants according to model-based evaluation using GPT-4o.}

\paragraph{Proof-reading and Paraphrasing.}

LLMs have been reported to provide useful assistance for scientific writing tasks, such as proof-reading, or providing suggestions for improving the writing style~\cite{salvagno2023can}. Additionally, some authors emphasize that LLMs can be helpful especially for non-native English speakers with regards to grammar, sentence structure, vocabulary and even translation, effectively serving as an English editing service~\cite{castellanos2023good,kim2023using}. Most papers on this topic are case studies, with results qualitatively evaluated by a human expert. 
\changed{\citet{jourdan2024casimir} introduce Casimir, a dataset of 15,646 OpenReview articles with sentence-level paired revisions. A later extension, ParaRev~\cite{jourdan2025pararev}, provides paragraph-level pairs, with a subset manually annotated with revision instructions. Experiments show that detailed instructions substantially improve automated revision quality under both statistical and model-based evaluations.}


\begin{changedpars}
\paragraph{Evaluation} Generated scientific texts are evaluated both with task-specific methods and those general to LLMs. 
Traditional reference-based metrics, such as BLEU~\cite{papineni2002bleu} or ROUGE~\cite{chin2004rouge},  
require associated reference output as a ground truth and have shown low correlation with human judgments~\cite{liu2016not}. With the rise of deep learning and LLMs, model-based approaches are gaining increasing importance~\cite{li2025generation,gao2025llm}. 
Although not exclusive to scientific texts, they focus on highly relevant dimensions such as coherence, consistency, and fluency~\cite{liu2023g,lee2025checkeval}. 
In a more domain-specific approach, \citet{huang2025papereval} propose \textbf{PaperEval}, a multi-agent system powered by LLMs to assess scientific paper quality across various dimensions (question, method, result, and conclusion). Its feasibility and effectiveness in distinguishing high-quality from low-quality papers have been tested on three evaluation datasets across four scientific fields (mathematics, physics, chemistry, and medicine). 
\end{changedpars}

\subsubsection{Domains of Application}

Text-based content generation is relevant for all scientific domains. \citet{liang2025quantifying} conduct a large-scale analysis across \changed{1M papers published between January 2020 and September 2024} to measure the prevalence of LLM-modified content over time. The papers they investigated were from a variety of disciplines and published on arXiv, bioRxiv, or Nature portfolio. 
Their results show the largest and fastest growth in computer science, with \changed{up to 22\% of the papers containing LLM-modified content; mathematics had the lowest prevalence (up to 9\%).}
According to arXiv's September 2024 Natural Language Learning \& Generation report~\cite{Leiter2024NLLGQA}, top-cited papers show notably fewer markers of AI-generated content as compared to random samples.

\begin{changedpars}
\subsubsection{Limitations and Future Directions}

LLMs and LLM applications demonstrate strong capabilities in tasks such as proofreading and paraphrasing, but still exhibit notable limitations for other tasks, making human-in-the-loop approaches essential. In particular, factual consistency, truthfulness, and bibliographic citations require human oversight. Rapid advances in LLMs further complicate evaluation, quickly rendering existing methods outdated and hindering reproducibility. Evaluating AI-generated text remains challenging: statistical metrics like BLEU and ROUGE lack semantic depth, model-based evaluations can be unreliable, and many tasks depend on small-scale human evaluations due the scarcity of appropriate benchmarks. Consequently, future research must prioritize robust benchmarks and datasets for scientific text generation. Beyond technical challenges, ethical concerns, including authorship, plagiarism, bias, and truthfulness, underscore the need for trustworthy and responsible AI systems.
\end{changedpars}


%% file: topics/multimodal.tex
\subsection{Multimodal Content Generation and Understanding}
\label{sec:multimodal}

\mybox{\textbf{Multimodal content generation} in the scientific domain refers to generating multimodal scientific content such as figures  
and tables in scientific papers, 
or slides and posters 
in a post-publication process. Automating such tasks via AI is important for 
multiple reasons: 
(i)~generating high-quality figures, tables, slides and posters is costly in terms of effort and time; 
(ii)~high-quality 
multimodal content in a paper 
can positively influence citation or acceptance decisions~\cite{Lee2016ViziometricsAV}; (iii)~tables, figures, posters, and slides make scientific content more accessible and compact.  
\textbf{Multimodal content understanding} refers to interpreting scientific images and tables, 
a prerequisite for answering questions about them or providing captions or summaries. These tasks are likewise effortful and time-consuming for human authors, pointing to a role for AI assistance.
}

\subsubsection{Data} In this section we detail datasets and benchmarks for selected multimodal tasks.  \changed{Further tasks, including table understanding and generation, can be found in Appendix~\ref{ax:multimodal_generation}, along with an overview table.}

\paragraph{Scientific Figure Understanding.} 
Scientific figure understanding benchmarks typically contain summaries or QA pairs for scientific figures.
\citet{10.1007/978-3-319-46493-0_15} provide a dataset with over 5K richly annotated diagrams and over 15K questions and answers. 
\textbf{FigureQA}~\cite{ebrahimi2018figureqa} is a synthetic dataset of over 100K scientific-style \mbox{(dot-)}line plots, vertical and
horizontal bar graphs, and pie charts, along with 1M associated questions 
generated using 15 different templates. 
Later research focuses on more challenging and realistic QA pairs. 
\textbf{ChartQA}~\cite{masry-etal-2022-chartqa}, for instance, provides complex reasoning questions concerning charts from various science-related sources. 
\textbf{CharXiv}~\cite{wang2024charxiv} 
is a manually curated dataset of descriptive and reasoning questions about 2.3K ``natural, challenging, and diverse'' charts from 
aXiv 
papers. \textbf{ArXivQA}~\cite{li-etal-2024-multimodal-arxiv}, a dataset of 35K scientific figures sourced from arXiv, contains 100K GPT-4o-generated, manually filtered QA pairs. 
\textbf{SPIQA}~\cite{pramanick2024spiqa} is a dataset of 270K manually and automatically created QA pairs that interpret complex scientific figures and tables. 
\citet{xu2024chartadapterlargevisionlanguagemodel} treat the problem of chart summarization with a 190K-instance dataset that builds on \textbf{ChartSumm}~\cite{Rahman2023ChartSummAC}, itself containing more than 84K charts along with their metadata and descriptions. 

\paragraph{Scientific Figure Generation.} Several datasets for scientific figure generation have been proposed. 
\changed{\textbf{DaTikZ}~\cite{belouadi2024automatikz,belouadi2024detikzify,belouadi2025tikzero} provides captions of scientific figures as instructions along with corresponding TikZ code, sourced from the \TeX\ Stack Exchange and arXiv submissions.
A later and larger version~\cite{greisinger2026tikzilla} improves the data quality through VLM-generated descriptions instead of noisy captions.}
For the task of converting sketches into scientific figures, the \textbf{SketchFig}~\cite{belouadi2024detikzify} benchmark provides 549 figure--sketch pairs sourced from the \TeX\ Stack Exchange. 
\changed{\textbf{DiagramGenBenchmark}~\cite{wei2025words} contains 6,713 training and 270 testing samples for diagram generation, and 1,400 training and 200 testing samples for diagram editing, sourced from freely licensed DOT and TikZ diagrams in VGQA and DaTikZ. \textbf{Plot2XML}~\cite{cui2025draw} includes 247 complex diagrams sourced from conference papers spanning multiple domains.}
\textbf{ChartMimic}~\cite{shi2024chartmimicevaluatinglmmscrossmodal} is a manually curated benchmark of 1,000 triplets of 
(instruction, code, figure)
instances for chart generation across various domains, including physics and economics. 
The data comes from human annotators writing Python code for prototype charts.  \textbf{ScImage}~\cite{zhang2024scimagegoodmultimodallarge} contains targeted template instructions focusing on different 
\changed{comprehension}
dimensions (spatial, numeric, attribute); 
for a subset of the data, the authors also provide reference figures that were
manually evaluated as being of high quality. In contrast to the other 
benchmarks, ScImage also contains instructions in languages other than English. 
\textbf{SciDoc2DiagramBench}~\cite{mondal-etal-2024-scidoc2diagrammer} is 
a benchmark comprised of 1,000 diagrams extracted from the presentation slides of 89 ACL papers, along with human-written ``intents''. 
The intents describe how the content from each paper can be translated into the diagrams for presentation purposes. 
\textbf{nvBench}~\cite{Luo2021nvBenchAL}, a benchmark of 25K tuples of natural language queries and corresponding visualizations, is drawn from 153 databases and contains more than 7K visualizations across seven chart types. nvBench is synthesized from natural language to SQL benchmarks.

\paragraph{Scientific Slide and Poster Generation.} 
Early efforts at automating the generation of presentation slides from scientific papers relied on relatively small datasets for development and evaluation. For example, \citet{Sravanthi2009SlidesGenAG} generate presentations from a modest collection of eight papers. Similarly, \citet{Hu2013PPSGenLT} and \citet{Wang2017PhraseBasedPS} use 1,200 and 175 paper--slide pairs, respectively. For scientific poster generation, \citet{Qiang2016LearningTG} 
collect
25 pairs of scientific papers and their corresponding posters.  
Such early datasets are often inaccessible due to various restrictions on distribution.

Two free-content datasets, \textbf{DOC2PPT}~\cite{Fu2021DOC2PPTAP} and \textbf{SciDuet}~\cite{sun-etal-2021-d2s}, have since emerged as widely used resources for scientific slide generation. The former consists of 5,873 scientific documents and their associated presentation slide decks, totalling around 100K slides, drawn from three research communities: computer vision, natural language processing, and machine learning. SciDuet 
has
1,088 papers and 10,034 slides from conferences such as ICML, NeurIPS, and the ACL Anthology. 
\changed{More recently, \textbf{SlidesBench}~\cite{ge2025autopresent} has provided 7K training and 585 test examples, each containing 20 slides on average, sourced from the web and covering ten broad domains (art, marketing, environment, technology, etc.).}

\subsubsection{Methods and Results} Here we survey approaches to multimodal content generation and understanding; a summary table, along with additional related works, is provided in Appendix~\ref{ax:multimodal_generation}.
 
\paragraph{Scientific Figure Understanding.} Scientific figure understanding is typically framed in terms of (visual) QA---e.g., whether models are able to adequately answer questions on a given input figure~\cite{10.1007/978-3-319-46493-0_15}. 
Several recent studies train baseline models such as transformers or relation networks~\cite{masry-etal-2022-chartqa,ebrahimi2018figureqa}, such as relation networks 
or apply recent LLMs to benchmarks~\cite{wang2024charxiv,li-etal-2024-multimodal-arxiv}. 
They generally  
show large gaps between proprietary models like GPT-4o and the strongest open models, and between all models and human performance. 
For chart summarization, \citet{Rahman2023ChartSummAC} find that older pre-trained language models such as BART and T5 suffer from hallucination and disregard important data points. \citet{xu2024chartadapterlargevisionlanguagemodel} propose ChartAdapter, a lightweight transformer module that can be combined with LLMs for improved modeling of chart summarization. 
\textbf{Evaluation} of scientific figure understanding benchmarks
tends to employ 
automatic metrics, many of which are now considered outdated or unreliable. For example, \citet{xu2024chartadapterlargevisionlanguagemodel} use BLEU and ROUGE. 
By contrast, \citet{pramanick2024spiqa} report both human and automatic evaluation, the latter including not just traditional QA metrics such as METEOR, ROUGE, and BERTScore, but also novel LLM-based metrics.

\paragraph{Scientific Figure Generation.}

Although work on automating visualization for science dates back to the 1980s at least, 
most recent work, including that covered here, involves multimodal LLMs. 
\changed{\textbf{AutomaTikZ}~\cite{belouadi2024automatikz} and \textbf{TikZero}~\cite{belouadi2025tikzero} fine-tune custom LLMs on DaTikZ for the text-to-TikZ task. \textbf{TikZilla}~\cite{greisinger2026tikzilla} extends this line of work by applying reinforcement learning using a custom image encoder for more semantically meaningful rewards. \textbf{VisCoder}~\cite{ni2025viscoder} uses Python libraries for visualization code generation. \textbf{DeTikZify}~\cite{belouadi2024detikzify} reconstructs scientific figures in TikZ from sketches or images. \textbf{Draw with Thought}~\cite{cui2025draw} uses multimodal LLMs for a preliminary coarse-to-fine planning approach, followed by structure-aware code generation with a self-refine mechanism for reconstruction. \textbf{DiagramAgent}~\cite{wei2025words} combines generation, reconstruction, and editing by introducing AI agents (plan, code, diagram-to-code, check).}
\citet{shi2024chartmimicevaluatinglmmscrossmodal}
aim to generate
Python code
from instructions and\slash or images, specifically focusing on charts.
They evaluate three proprietary and 11 open-weight LLMs on their ChartMimic benchmark, finding that even the best models (GPT-4 and Claude-3-opus) have substantial room for improvement. 
\citet{zhang2024scimagegoodmultimodallarge} provide a template-based approach to evaluate various multimodal LLMs in generating scientific images. 
They explore those that generate TikZ and Python code for images, as well as those that generate images directly,\footnote{\changed{Most scientific figure generation approaches are based on code generation, while direct image generation approaches are rare. The likely reason is that code generation methods usually work better (e.g., allowing higher precision when plotting numeric inputs) and their output is easier for users to post-edit. Within code generation, those employing general-purpose languages like Python seem to perform better than those targeting specialized drawing languages like TikZ, though current comparisons are restricted to individual use cases~\cite{zhang2024scimagegoodmultimodallarge}.}} and also consider different input languages (English, German, Chinese, Farsi). They find that, except for GPT-4o, most models struggle substantially in generating adequate scientific images. 
\citet{Zala2023DiagrammerGPT} 
explore the diagram generation task where LLMs first generate diagram plans and then the diagrams themselves. 
\citet{mondal-etal-2024-scidoc2diagrammer} explore the same task with an additional refinement---namely, feedback from multiple critic models---to enhance factual correctness. 
\textbf{Evaluation} uses automatic metrics including DreamSim~\cite{10.5555/3666122.3668330} for image similarity, Crystal BLEU~\cite{10.1145/3551349.3556903} for code similarity, and CLIPScore~\cite{hessel-etal-2021-clipscore} for text--image similarity, as well as manual evaluation by domain experts. The former are typically reported to have low or medium correlation with the latter, establishing the need for domain-specific evaluation in future work.

\paragraph{Scientific Slide and Poster Generation.}

Early work on scientific slide generation~\cite{Sravanthi2009SlidesGenAG,Hu2013PPSGenLT,Wang2017PhraseBasedPS, Li2021TowardsTS} used extractive approaches, copying text from papers to serve as slide content. 
Later systems explored abstractive approaches to generate the textual content of slides, such as with sequence-to-sequence models~\cite{Fu2021DOC2PPTAP,sun-etal-2021-d2s}.
Researchers are now exploring
(multimodal) LLMs 
for generating slides using natural language prompts~\cite{mondal-etal-2024-presentations,maheshwari-etal-2024-presentations,bandyopadhyay-etal-2024-enhancing-presentation1}. 
\changed{AutoPresent~\cite{ge2025autopresent}, for example, fine-tunes an LLM using SlidesBench to generate slides from detailed natural language instructions with images, detailed instructions only, or high-level instructions.}
However, modern systems still take an extractive approach to multimodal content, copying images or tables directly from the original papers instead of generating new ones~\cite{Sravanthi2009SlidesGenAG,sun-etal-2021-d2s,Fu2021DOC2PPTAP,mondal-etal-2024-presentations,bandyopadhyay-etal-2024-enhancing-presentation1}.
The task of poster generation has received less attention, with studies mainly exploring ML-based methods for generating key content and panel layouts~\cite{Qiang2016LearningTG,Xu2022PosterBotAS}.
\textbf{Evaluation} of slide generation has involved both human judgments and automatic metrics (most commonly ROUGE).  
\citet{Fu2021DOC2PPTAP} introduce some novel metrics: Longest Common Figure Subsequence, which measures the quality of figures in the generated slides; Text--Figure Relevance, which assesses the similarity between the text of the ground truth slide and the generated slide containing the same figure; and Mean Intersection over Union, which evaluates layout quality. Recent studies have also used LLMs to assess the quality of generated slides~\cite{bandyopadhyay-etal-2024-enhancing-presentation1,maheshwari-etal-2024-presentations}. For scientific poster generation, in addition to conducting user studies, \citet{Qiang2016LearningTG} measure the mean squared error (MSE) of the panel parameters such as panel size and aspect ratio.

\subsubsection{Domains of Application}

Many recent datasets for multimodal content generation and understanding are drawn from arXiv and more generally the STEM domain. Models such as DeTikZify and AutomaTikZ have also been fine-tuned on such data. This indicates a limitation both in terms of application scenarios and model assessments, as these may perform worse when applied in cross-domain scenarios. 

\subsubsection{Limitations and Future Directions}

Limitations common to the approaches we have discussed include (1)~the comparatively small datasets for fine-tuning models; (2)~sub--human-level performance on recent benchmarks, particularly for non-proprietary models; (3)~over-representation of arXiv and STEM domains in training and evaluation; (4)~models' lack of reasoning abilities; and (5)~lack of reliable, task-specific evaluation methods, particularly for generative tasks. Some problems are task-specific: for example, for table generation, the input text may be very long, which constitutes a problem for many current LLMs; for slide generation, there are no approaches that can generate slides from multiple documents (e.g., for tutorials) or that generate content beyond that contained in a reference paper (which may be necessary for including relevant background material); for figure generation, models like AutomaTikZ are trained on captions, which are often not appropriate for generating the corresponding figure (e.g., a caption may simply be ``Proof of Theorem X''). \changed{Reproducibility is hampered by some studies' use of datasets that are not freely distributable---for example, one study~\cite{greisinger2026tikzilla} reports that only a third of all arXiv-harvested papers are permissively licensed.}


%% file: topics/peerreview.tex
\subsection{Peer Review}
\label{sec:peer_review}

\begin{figure*}
  \centering
  \includegraphics[width=\textwidth]{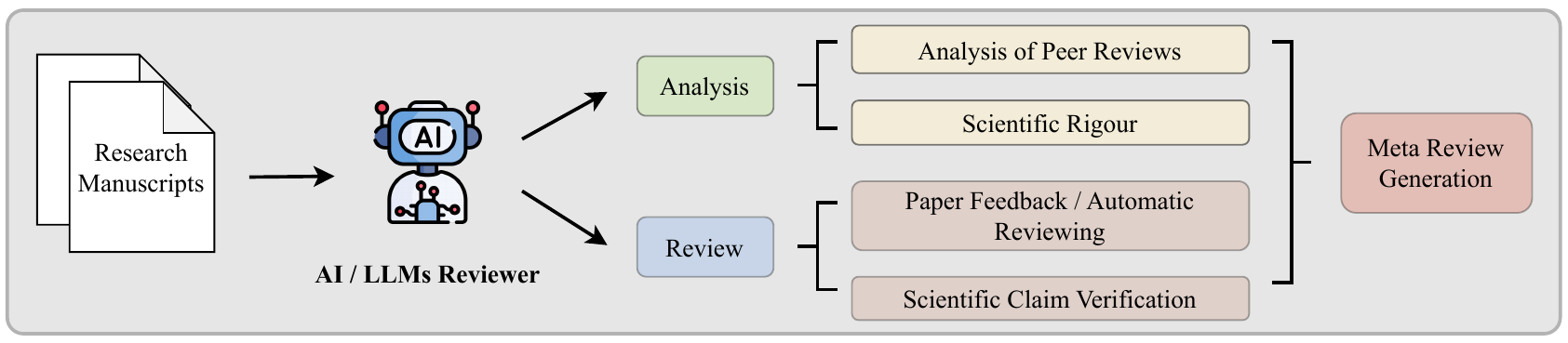}
  \caption{Process of AI-enhanced peer review. In the analysis step, the LLM reviewer examines research manuscripts and evaluates peer reviews to assess scientific rigor. The review step involves providing feedback on the paper and verifying scientific claims. Finally, the gathered information is synthesized to generate a final meta-review.} 
  \label{fig:peer_review_overview}
\end{figure*}

\mybox{The highest standard in scientific quality control is \textbf{peer reviewing}. In this process, authors present their scientific arguments (e.g., the findings of a study, or a grant proposal) in form of a manuscript to their peers, who then assess its scientific validity and quality. Often this process has multiple stages, as shown in Fig.~\ref{fig:peer_review_overview}. For instance, in the ACL Rolling Review (ARR) system, 
reviewers write detailed assessments whose arguments and questions the authors may then rebut and clarify to convince the reviewers to raise their scores. A meta-reviewer then evaluates this discussion and submits to the program chairs an acceptance\slash rejection recommendation, which may or may not be adopted. During this process, multiple (potentially multi-modal) artifacts are processed and created---mainly the manuscript under review, the written reviews, the author--reviewer discussion, and the meta-review. 
In general, peer review is considered a challenging,  subjective process, where reviewers are prone to unfair biases like sexism and racism, and often rely on expedient heuristics~\cite{strauss2023racism,regner2019committees}. In some fields, these problems are compounded by an exploding number of submissions~\cite{kunzli2022not}, pushing review systems to their limits.
To counteract this situation, researchers have addressed various several problems under the umbrella of AI-supported peer review. Related overviews on the topic, or on some of its aspects,
point to its importance and timeliness~\cite{kousha,drori2024human,staudinger-etal-2024-analysis,lin2023automated,checco,kuznetsov2024can}.  
Here, we focus on approaches to the most established tasks related to peer review, following the same structure as in previous sections.}

\subsubsection{Data}
Peer reviewing data is scarce: few scientific communities publish reviewing artifacts at all, let alone under permissive licenses. 
Exceptions include PeerRead~\cite{kang-etal-2018-dataset}, which collects review data from various sources (e.g., ACL, ICRL), and CiteTracked~\cite{Plank2019CiteTrackedAL}, which also contains citation information.
NLPeer~\cite{dycke-etal-2023-nlpeer}, a model for how larger-scale open publishing of raw peer reviewing data could work, uses a corpus of ARR reviews where the consent of the respective actors was explicitly obtained. 
\begin{table}
\small
\caption{Annotated and\slash or task-specific datasets for analyzing peer reviewing.}
\label{tab:data_peer_reviewing}
\centering
\begin{tabular}{lp{2.9cm}lp{52mm}}
\toprule
\textbf{Dataset}  & \textbf{Size} & \textbf{Sources} & \textbf{Application} \\
\midrule
HedgePeer~\cite{10.1145/3529372.3533300} & 2,966 documents & ICLR 2018 & Uncertainty detection\\
PolitePeer~\cite{politepeer} &2,500 sentences & ICLR et al. & Politeness analysis\\
COMPARE~\cite{singh2021compare} & 1,800 sentences & ICLR & Comparison analysis\\
ReAct~\cite{Choudhary_2021} & 1,250 comments & ICLR & Actionability analysis\\
MReD~\cite{shen-etal-2022-mred} & 7,089 meta-reviews & ICLR & Meta-review analysis and generation\\ 
CiteTracked~\cite{Plank2019CiteTrackedAL} & 3,427 papers, 12K reviews & NeurIPS & Citation prediction \\
MOPRD~\cite{Lin_2023} & 6,578 papers & PeerJ & Review comment generation \\ 
Revise and Resubmit~\cite{10.1162/coli_a_00455} & 5.4K papers & F1000Research & Tagging, linking, version alignment  \\
ORB~\cite{szumega2023open} & 92,879 reviews & OpenReview, SciPost & Acceptance prediction\\ 
ARIES~\cite{d2023aries} & 3.9K comments & OpenReview & Feedback--edits alignment, revision generation  \\ 
DISAPERE~\cite{kennard-etal-2022-disapere} & 506 review--rebuttal pairs & ICLR & Review action analysis, polarity prediction, review aspect \\
PeerReviewAnalyze~\cite{10.1371/journal.pone.0259238} & 1,199 reviews & ICLR & Review paper section correspondence, paper aspect category detection, review statement role prediction, review statement significance detection, meta-review generation \\
JitsuPeer~\cite{purkayastha-etal-2023-exploring} & 9,946 review and 11,103 rebuttal sentences & ICLR & argumentation analysis, canonical rebuttal scoring, review description generation, end2end canonical rebuttal generation\\
\bottomrule
\end{tabular}
\end{table}
Several annotated datasets support tasks in peer review analysis, an overview of which is provided in Table~\ref{tab:data_peer_reviewing}. 
Recent curated resources have focused on complex tasks such as understanding the effect of peer review feedback on manuscript revisions~\cite{d2023aries} or identifying the attitudes underlying specific criticisms in reviews~\cite{purkayastha-etal-2023-exploring}.

\subsubsection{Methods and Results}

\changed{Pre-LLM approaches~\cite[e.g.,][]{10.1145/3529372.3533300}} were mostly based on traditional ML methods, targeting simpler analyses involving sentence classification tasks. Later, deep learning approaches~\changed{\cite[e.g.,][]{hua-etal-2019-argument}} (including those based on pre-trained language models) and more complex analyses (e.g., of argumentation) defined the state of the art in computational peer review processing. Researchers have now started exploring 
LLMs 
in prompting-based frameworks, for complex tasks like peer review generation and meta-review generation~\changed{\cite[e.g.,][]{liu2023reviewergpt}. Below we provide an overview of the methods and results for the most common tasks under the umbrella of peer reviewing. Information on related tasks, such as scientific claim verification, can be found in Appendix~\ref{ax:peerreview}.}

\paragraph{Analysis of Peer Reviews}

Prior 
works have analyzed peer reviews for a multitude of aspects, like uncertainty~\cite{10.1145/3529372.3533300}, politeness~\cite{politepeer}, and sentiment~\cite{Chakraborty_2020}. 
However, given that science as a whole and especially peer review relies to a large extent on convincing 
peers, large efforts have 
been spent on understanding arguments or argument-related aspects (e.g., substantiation of arguments) in peer review artifacts~\cite[e.g.,][]{Fromm2021,hua-etal-2019-argument}.
Here, most approaches have used pre-trained language models. For instance, \citet{hua-etal-2019-argument} work on mining the arguments in peer reviews using conditional random fields, 
LSTMs, 
and BERT. In contrast, \citet{guo-etal-2023-automatic} and \citet{Fromm2021} fully rely on (domain-adjusted) pre-trained language models for argument mining: SciBERT, ArgBERT, and PeerBERT. \citet{cheng-etal-2020-ape} leverage multi-task learning approaches based on LSTMs and BERT. In a similar vein, \citet{purkayastha-etal-2023-exploring} study the generation of rebuttals for author--reviewer discussions based on jiu-jitsu argumentation, a specific argumentation theory. 

\paragraph{Paper Feedback and Automatic Reviewing}

Several works have explored methods to provide general feedback on scientific publications to fully or partially automate peer reviews.  
For instance, \citet{li-etal-2020-multi-task} propose a multi-task learning approach for peer review score prediction, where different aspect score prediction tasks (e.g., novelty) can inform each other. \citet{ghosal-etal-2019-deepsentipeer} leverage the concept of sentiment to predict scores based on review texts. In a similar vein, \citet{10.1007/978-3-030-91669-5_33} leverage paper--review interactions to predict final decisions of a review process. 
\citet{wang-etal-2020-reviewrobot} focus on explainability during review score prediction for several review categories by constructing knowledge graphs (e.g., representing the background of a paper). More recent works have included generation of feedback texts into the problem setup.
\citet{bartoli2020} frame the problem as exploring the potential of GPT-2 for conducting academic fraud by generating fake reviews. 
In contrast, \citet{10.1613/jair.1.12862} ask whether it would be possible to automate reviewing leveraging 
targeted summarization models, a recently trending topic. 
\citet{liu2023reviewergpt} explore prompting-based review generation with various LLMs, 
finding that GPT-4 performs best and that task granularity matters. \citet{robertson2023gpt4} similarly finds GPT-4 to be ``slightly'' helpful for peer reviewing, and \citet{liang2023can} demonstrate in a comparative study that users of a GPT-4--based peer review system found the feedback to be (very) helpful more than half the time.
\citet{d2024marg} show a multi-agent approach with LLMs that engage in a discussion to produce better results than a single model.

\paragraph{Meta Review Generation}

\citet{9651825} tackle meta-review generation using a multi-encoder transformer network, and \citet{li-etal-2023-summarizing} use a multi-task learning approach for refining pre-trained language models for the task. \citet{stappen2020uncertainty} explore the aggregation of reviews for providing additional computational decision support to editors based on uncertainty-aware methods like soft labeling. Both \citet{zeng2023meta} and \citet{santu2024prompting} rely on 
LLMs, 
which they specifically prompt for the task. \changed{In contrast, \citet{purkayastha2025decision} propose a conversational agent that interactively supports meta-reviewers in their decision-making.}

\subsubsection{Domains of Application}

Peer review setups vary in the aspects to review for, scoring schemes, expected review length, and the stages and dynamics of the reviewer--author and reviewer--reviewer discussions. Thus, while none of the studies presented above targets a problem unique to any scientific discipline, the particularities will likely be very different for each specific community and existing systems must be evaluated or adapted before deployment.

\subsubsection{Limitations and Future Directions}

The variety of scientific domains whose peer review has been studied is still limited. Most work relies on data from OpenReview, a platform used primarily by the representation learning and NLP communities; 
other disciplines may be wholly unrepresented in existing peer review datasets.  \changed{Even within communities, there can be great variance in how peer reviews are conducted, which limits the comparability of approaches.}
Scientific rigor in particular remains unexplored, despite being a critical aspect of peer review; most existing studies rely on predefined rigor checklists 
that are not easily scalable or transferable across domains.
Given these gaps, future research could benefit from exploring new domains of peer review, developing domain adaptation approaches, and advancing models for assessing scientific rigor. \changed{Reproducibility studies and larger benchmarks could further advance the field.}  Additionally, 
ethical concerns demand the prioritization of
research on \emph{trustworthy} AI support for peer review, ensuring that human experts retain autonomy in the process. 


%% file: ethics.tex
\section{Ethical Concerns}\label{sec:ethics}

There is by now a growing body of work addressing major ethical concerns related to generative AI. 
\citet{baldassarre2023social}, for instance, present a systematic literature review regarding the social impact of generative AI, especially taking into account 71 papers on ChatGPT. They identify privacy, inequality, bias, discrimination, and stereotypes as areas of concern. Another literature review on ethics and generative AI~\cite{hagendorff2024mapping} identifies jailbreaking, hallucination, alignment, harmful content, copyright, private data leakage, and impacts on human creativity as topics of increasing interest. This review furthermore identifies 19 distinct clusters of ethics topics, with fairness\slash bias being the most frequently mentioned, followed by safety, harmful content\slash toxicity, hallucination, privacy, interaction risks, security\slash robustness on ranks two to six, with writing\slash research on rank 18.
\citet{ali2024ethical} review 364 recent papers on generative AI and ethics published from 2022 to 2024 in different domains including the use of generative AI in scientific research. Topics identified as critical to academia are authenticity, intellectual property, and academic integrity. 
\citet{sun2024trustllm} argue that in application areas such as scientific research, ensuring the trustworthiness of LLMs is crucial. \changed{\citet{dergaa2023moving} find the use of AI chatbots in academic research heavily associated with stigma and propose mitigation strategies.}

Truthfulness---i.e., the accurate representation of information, facts and results---is a particularly essential challenge for LLMs.  Benchmarks and datasets developed to evaluate different aspects of it include TruthfulQA~\cite{lin2021truthfulqa}, HaluEval~\cite{li2023halueval}, and FELM~\cite{zhao2024felm}, for identifying hallucinations; SelfAware~\cite{yin2023large} for assessing awareness of knowledge limitations; FreshQA~\cite{vu2023freshllms} and Pinocchio~\cite{yin2023large} for exploring adaptability to rapidly evolving information; and
TrustLLM~\cite{sun2024trustllm}, which incorporates existing and new datasets not just on truthfulness but also safety, fairness, robustness, privacy, and machine ethics. Evaluations with TrustLLM show that proprietary LLMs generally outperform open-source LLMs in trustworthiness, 
Llama2~\cite{touvron2023llama} being a notable exception. 
However, proprietary LLMs (including Llama2) often struggle to provide truthful responses when relying solely on internal knowledge.  Their performance does improve significantly with additional external knowledge, and there exists a positive correlation between trustworthiness and the functional effectiveness of the model in 
downstream
tasks. 

Editors of scientific publications are particularly challenged by the increasing proportion of AI-generated text in manuscripts~\cite{gray2024chatgptcontaminationestimatingprevalence, kobak2024delving, liang2024mapping, cheng2024have} and by its potential use in peer reviewing (cf.~§\ref{sec:peer_review}). The editors of the \emph{Journal of Information Technology} have elaborated on the limitations and risks of using generative AI in the production of scientific publications~\cite{schlagwein2023chatgpt}, referring to an eight-point ``Artificial Imperfections'' test to illustrate current limitations of generative AI:
AI is (1)~brittle, (2)~opaque, (3)~greedy, (4)~shallow and tone-deaf, (5)~manipulative and hackable, (6)~biased, (7)~invasive, (8)~``faking it''~\cite[p.\,107]{willcocks2023maximizing}.
Nevertheless, they conclude that although AI should not be forbidden, authors must take full responsibility for its output and adhere to the ``scientific principle of transparency'' by giving full and transparent disclosure of their usage of AI, and moreover that ``it is then up to the reviewers and editors to assess and make decisions on the specific use of that generative AI in a specific piece of research.''
Guidelines proposed by the editors of \emph{iMeta} similarly hold authors fully responsible for the integrity of their manuscripts, and for addressing ethical concerns and ensuring the accuracy and fairness of AI-generated content, complying with data protection and privacy laws, and considering the relevant copyright and intellectual property issues~\cite{pu2024chatgpt}.  They furthermore state that AI‐assisted technologies cannot be recognized as authors, that the use of generative AI must be transparently disclosed (including the prompts and specific versions of the tools used), that AI-generated images and multimedia should be accepted only when specifically allowed, and that the use of AI in the reviewing process is expressly prohibited.


%% file: conclusion.tex
\section{Conclusion}\label{sec:conclusion}

In this paper, we surveyed approaches in the area of AI4Science, with a particular focus on recent large language model-based methods. We examined five key aspects of the research cycle: (1)~search, (2)~experimentation and research idea generation, (3)~text-based content production, (4)~multimodal content production, and (5)~peer review. For each topic, we discussed relevant datasets, methods, and results, including evaluation strategies, while highlighting limitations and avenues for future research. Ethical concerns featured prominently in our survey, given the potential for misuse and challenges in maintaining scientific integrity in the face of AI-assisted content generation.

\changed{Overall, while recent advances suggest that AI systems can meaningfully support certain components of the scientific workflow, their current capabilities remain limited and uneven. Many methods rely on narrow benchmarks, struggle with generalization, or require substantial human oversight to avoid errors, bias, or misinterpretation. Consequently, AI4Science should presently be viewed as a complementary set of tools rather than a transformative replacement for human expertise.} We hope that this survey inspires new initiatives in AI4Science, driving faster, more efficient, and more inclusive scientific discovery, experimentation, reporting and content synthesis---while upholding the highest ethical standards. As the ultimate goal of science is to serve humanity, we hope these advancements will accelerate knowledge creation and enhance the accessibility and reliability of research, leading to improved healthcare, medical treatments, economic processes, among a myriad of other societal benefits.


%% file: appendix.tex
\appendix

\section{Historical Context and Background}\label{ax:background}

\begin{figure}[b]
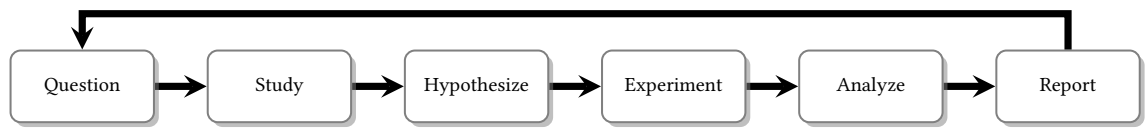

  \centering
\begin{adjustbox}{width = \textwidth}%
  \smartdiagramset{uniform color list=white!20 for 6 items, arrow color=black, uniform arrow color=true}%
  \smartdiagram[flow diagram:horizontal]{Question,Study,Hypothesize,Experiment,Analyze,Report}%
\end{adjustbox}
  \caption{Scientific discovery cycle, after~\cite{cornelio2023combining}}%
  \label{fig:cycle}%
\end{figure}
 
Throughout history, science has undergone a number of paradigm shifts, culminating in today's era of data-intensive exploration~\cite{hey2009jim}.  Although new tools and frameworks have accelerated the pace of scientific discovery, its basic steps have remained unchanged for centuries.  As visualized in Fig.~\ref{fig:cycle}, these include (1)~conception of a research question or problem, typically arising from a gap in disseminated knowledge; (2)~collection and study of existing literature or data relevant to the problem; (3)~formulation of a falsifiable hypothesis; (4)~design and execution of experiments to test this hypothesis; (5)~analysis and interpretation of the resulting data; and (6)~reporting on the findings, allowing for their exploitation in real-world applications or as a source of knowledge for a further iteration of the scientific cycle.

With respect to the first two of these steps, a major challenge for any scholar is achieving, and then maintaining, sufficient familiarity with existing research on a given topic to be able to identify new research questions or to discover the knowledge required to answer them.  Before the 20th century, it was often feasible to keep abreast of developments in a specialty simply by reading all the relevant books and journals as they were published.  In modern times, however, the number of scientific publications has been doubling every 17 years~\cite{bornmann2021growth}, making this exhaustive approach unworkable.  The need to sift through large quantities of scholarly knowledge spurred the specialization of simple library catalogs (in use since ancient times) into abstracting journals, bibliographic indexes, and citation indexes.  By the 1960s and 1970s, many of these resources were being produced with standardized control principles and technologies, and could be queried interactively using automated information retrieval systems~\cite[pp.\,88--91]{borgman2007scholarship}.  These technical developments have enabled the widespread adoption of more principled approaches to the exploration of scientific knowledge, such systematic reviews~\cite{chalmers2002brief} and citation analysis~\cite{garfield1955citation}.

How experts propose hypotheses to explain observed phenomena has been extensively discussed in the philosophy and psychology of science, albeit with little empirical work until relatively recently~\cite{clement1989learning,clement2022multiple}.  Contrary to the idealized notion of scientific reasoning, hypotheses rarely come about solely through induction (i.e., the abstraction of a general principle from a set of empirical observations).  Rather, case studies employing think-aloud protocols suggest that hypotheses are generated through a process of successive refinement.  These processes may involve non-inductive heuristics (analogies, simplifications, imagistic reasoning, etc.)\ that often fail individually, but may lead to valid explanatory models after ``repeated cycles of generation, evaluation, and modification or rejection''~\cite{clement1989learning,clement2022multiple}.

Experimentation and analysis aim to establish a causal relationship between the independent and dependent variables germane to a given scientific hypothesis.  The metascientific literature abounds with practical advice on the design and execution of experiments, much of it discipline-specific. However, the general ideas at play can be traced to Ronald Fisher, whose seminal works on statistical methods~\cite{fisher1925statistical} and experimental design~\cite{fisher1935design} popularized the principles of randomization (assigning experimental subjects by chance), replication (observing different experimental subjects under the same conditions), and blocking (eliminating undesired sources of variation).  Besides these considerations, experimental design involves the determination of the (statistical) analysis that will be performed, and is often constrained by the availability of resources such as the time, effort, or cost to gather and analyze observations or data~\cite{kirk2009experimental}.

The final step in the scientific cycle, reporting, encompasses the dissemination of research findings, typically but not exclusively to the wider scientific community through articles, books, and presentations.  The practice of scientific communication has itself attracted scientific study, leading to descriptive and pedagogical treatments of its various processes and strategies \cite[e.g.,][]{yore2004scientists,hartley2008academic}.  The essential role of peer review~\cite{weller2001editorial} has attracted special attention, albeit more on its high-level processes, its efficacy and reliability, and its objectivity and bias rather than on how reviewers go about evaluating manuscripts and communicating this evaluation.  Accordingly, technological developments in the peer review workflow have until very recently tended to focus on managing or streamlining the review process for the benefit of the editor and publisher, or on supporting open or collaborative reviewing~\cite{weller2001editorial,drozdz2024peer}.

\section{Supplementary Material on AI Support for Specific Topics and Tasks}

\subsection{Literature Search, Summarization, and Comparison}
\label{ax:search_engine}

\paragraph{Search Engines}

\begin{table}
\caption{Overview of additional literature search engines and benchmarks.  $\checkmark$  indicates feature availability; empty cells indicate lack of features or publicly documented support.}
\label{tab:ax_google_scholar}
\vspace{1.5cm}
\centering
\resizebox{\columnwidth}{!}{%
\begin{tabular}{lllll|llll|lllll|lllll|ll}
\multicolumn{1}{c}{\textbf{}} & \multicolumn{1}{c}{\textbf{Platform}} & \multicolumn{1}{c}{\rotatebox[origin=l]{45}{\makebox[0pt][l]{\textbf{Search}}}} & \multicolumn{1}{c}{\rotatebox[origin=l]{45}{\makebox[0pt][l]{\textbf{Recommendations}}}} & \multicolumn{1}{c}{\rotatebox[origin=l]{45}{\makebox[0pt][l]{\textbf{Collections}}}} & \multicolumn{1}{c}{\rotatebox[origin=l]{45}{\makebox[0pt][l]{\textbf{Citation Analysis}}}} & \multicolumn{1}{c}{\rotatebox[origin=l]{45}{\makebox[0pt][l]{\textbf{Trending Analysis}}}} & \multicolumn{1}{c}{\rotatebox[origin=l]{45}{\makebox[0pt][l]{\textbf{Author Profiles}}}} & \multicolumn{1}{c}{\rotatebox[origin=l]{45}{\makebox[0pt][l]{\textbf{Visualization Tools}}}} & \multicolumn{1}{c}{\rotatebox[origin=l]{45}{\makebox[0pt][l]{\textbf{Paper Chat}}}} & \multicolumn{1}{c}{\rotatebox[origin=l]{45}{\makebox[0pt][l]{\textbf{Idea Generation}}}} & \multicolumn{1}{c}{\rotatebox[origin=l]{45}{\makebox[0pt][l]{\textbf{Paper Writing}}}} & \multicolumn{1}{c}{\rotatebox[origin=l]{45}{\makebox[0pt][l]{\textbf{Summarization}}}} & \multicolumn{1}{c}{\rotatebox[origin=l]{45}{\makebox[0pt][l]{\textbf{Paper Review}}}} & \multicolumn{1}{c}{\rotatebox[origin=l]{45}{\makebox[0pt][l]{\textbf{Datasets}}}} & \multicolumn{1}{c}{\rotatebox[origin=l]{45}{\makebox[0pt][l]{\textbf{Code Repositories}}}} & \multicolumn{1}{c}{\rotatebox[origin=l]{45}{\makebox[0pt][l]{\textbf{LLM Integration}}}} & \multicolumn{1}{c}{\rotatebox[origin=l]{45}{\makebox[0pt][l]{\textbf{Web API}}}} & \multicolumn{1}{c}{\rotatebox[origin=l]{45}{\makebox[0pt][l]{\textbf{Personalization}}}} & \multicolumn{1}{c}{\textbf{Cost}} & \multicolumn{1}{c}{\textbf{Data Size}} \\ \midrule
\multirow{12}{*}{\centering \rotatebox[origin=c]{90}{\textbf{Search Engines}}} & \href{https://scholar.google.com}{Google Scholar} & $\checkmark$ & $\checkmark$ & $\checkmark$ & $\checkmark$ &  & $\checkmark$ &  &  &  &  &  &  &  &  &  &  & $\checkmark$ & Free &  \\ 
 & \href{https://www.semanticscholar.org}{Semantic Scholar} & $\checkmark$ & $\checkmark$ & $\checkmark$ & $\checkmark$ & $\checkmark$ & $\checkmark$ &  & $\checkmark$ &  &  & $\checkmark$ &  &  &  & $\checkmark$ & $\checkmark$ & $\checkmark$ & Free & 214M \\ 
 & \href{https://xueshu.baidu.com}{Baidu Scholar} & $\checkmark$ & $\checkmark$ & $\checkmark$ & $\checkmark$ & $\checkmark$ & $\checkmark$ &  &  &  &  &  &  &  &  & $\checkmark$ &  & $\checkmark$ & Freemium & 680M \\ 
 & \href{https://www.base-search.net}{BASE} & $\checkmark$ &  & $\checkmark$ &  &  &  &  &  &  &  &  &  &  &  &  & $\checkmark$ &  & Free & 415M \\ 
 & \href{https://scholar.archive.org/}{Internet Archive Scholar} & $\checkmark$ &  &  &  &  &  &  &  &  &  &  &  &  &  &  & $\checkmark$ &  & Free & 35M \\ 
 & \href{https://www.scilit.net}{Scilit} & $\checkmark$ &  & $\checkmark$ & $\checkmark$ &  & $\checkmark$ &  &  &  &  &  &  &  &  &  &  &  & Free & 172M \\ 
 & \href{https://www.lens.org/}{The Lens} & $\checkmark$ &  & $\checkmark$ &  &  & $\checkmark$ &  &  &  &  &  &  &  &  &  & $\checkmark$ &  & Freemium & 284M \\ 
 & \href{https://science.gov}{Science.gov} & $\checkmark$ &  &  &  &  &  & $\checkmark$ &  &  &  &  &  &  &  &  &  &  & Free & 200M \\ 
 & \href{https://www.academia.edu/}{Academia.eu} & $\checkmark$ &  & $\checkmark$ &  &  & $\checkmark$ &  &  &  &  &  &  &  &  &  &  &  & Freemium & 55M \\ 
 & \href{https://openalex.org/}{OpenAlex} & $\checkmark$ &  &  &  &  & $\checkmark$ &  &  &  &  &  &  &  &  &  & $\checkmark$ &  & Freemium &  \\ 
 & \href{https://www.acemap.info/}{AceMap} & $\checkmark$ &  &  & $\checkmark$ & $\checkmark$ & $\checkmark$ & $\checkmark$ &  &  &  &  &  & $\checkmark$ &  &  &  &  & Free & 260M \\ 
 & \href{https://www.ncbi.nlm.nih.gov/research/pubtator3/tutorial}{PubTator3} & $\checkmark$ &  & $\checkmark$ & $\checkmark$ &  &  &  &  &  &  &  &  &  &  &  & $\checkmark$ &  & Free & 6M \\ \midrule
\multirow{4}{*}{\centering \rotatebox[origin=c]{90}{\textbf{Benchm.}}} & \href{https://portal.paperswithcode.com/}{Papers with Code} & $\checkmark$ &  &  &  &  &  &  &  &  &  &  &  & $\checkmark$ & $\checkmark$ &  &  &  & Free & 154K \\ 
 & \href{https://github.com/OSU-NLP-Group/ScienceAgentBench}{ScienceAgentBench} &  &  &  &  &  &  &  &  &  &  & $\checkmark$ &  & $\checkmark$ & $\checkmark$ & $\checkmark$ &  &  & Free &  \\ 
 & \href{https://orkg.org/benchmarks}{ORKG Benchmarks} &  &  &  &  & $\checkmark$ &  & $\checkmark$ &  &  &  &  &  & $\checkmark$ &  &  &  &  & Free &  \\ 
 & \href{https://huggingface.co/}{Huggingface} & $\checkmark$ &  & $\checkmark$ &  & $\checkmark$ &  &  &  &  &  &  &  & $\checkmark$ & $\checkmark$ &  &  &  & Freemium &  \\ 
\end{tabular}%
}
\end{table}

\begin{changedpars}
Traditional academic search engines such as \href{https://scholar.google.com}{Google Scholar}, \href{https://www.semanticscholar.org}{Semantic Scholar}, \href{https://xueshu.baidu.com}{Baidu Scholar}, \href{https://science.gov}{Science.gov}, and \href{https://www.base-search.net}{BASE}, as shown in Table~\ref{tab:ax_google_scholar}, are characterized by their broad literature coverage, citation tracking capabilities, and keyword-based search functionality. Their primary advantages include extensive indexing of scholarly content, which involves aggregating and organizing vast amounts of academic documents from various sources such as publisher websites, institutional repositories, and open-access archives. This comprehensive indexing spans multiple disciplines and document types, ensuring that users can access a diverse set of resources. Additionally, these platforms offer citation analysis features that allow researchers to track citation counts, measure the impact of publications, and explore citation networks to identify influential works and emerging trends within a given field. Another significant advantage is their free access to a wide range of academic resources, such as peer-reviewed journal articles, conference papers, preprints, theses and dissertations, technical reports, books and book chapters, as well as grey literature like white papers, government reports, and institutional research outputs. However, these search engines have certain limitations, such as limited filtering options and relatively basic relevance ranking mechanisms compared to more advanced AI-enhanced search tools.
\end{changedpars}

\paragraph{Benchmarks and Leaderboards}

Code and dataset-focused search engines include platforms such as \href{https://huggingface.co/}{Huggingface} (which in 2025 absorbed Papers with Code) and \href{https://github.com/OSU-NLP-Group/ScienceAgentBench}{ScienceAgentBench}, which are specifically designed to bridge the gap between academic publications and practical implementation by linking research papers with associated code and datasets. These platforms facilitate reproducibility and practical application of research findings by aggregating code repositories, enabling researchers and practitioners to easily explore implementations, compare results, and benchmark their models. A key feature of such platforms is their provision of dataset discovery tools, which allow users to identify relevant datasets for specific research problems, fostering collaboration, and accelerating experimentation cycles. These search engines are particularly valuable for machine learning practitioners, as they facilitate quick access to ready-to-use codebases, helping them implement cutting-edge research more efficiently. Based on these community-curated leaderboards, some studies have proposed models for constructing leaderboards directly from scientific papers~\cite{hou-etal-2019-identification,kardas-etal-2020-axcell,sahinuc-etal-2024-efficient}. 

\begin{movedpars}
\paragraph{Ethical Concerns}
The use of AI in scientific search, summarization, and comparison raises ethical considerations, particularly in ensuring transparency, accountability, and equity. AI can significantly accelerate the pace of discovery, automate search tasks, and uncover patterns that may elude human researchers, but it also introduces risks and biases. Existing dynamics such as the Matthew effect, where well-known researchers receive disproportionate attention, might be algorithmically reinforced, intensifying inequalities. We believe that research should follow a human-centric approach, in which the human researcher is provided with advanced tools but remains fully responsible for executing the research and summarizing the results in research papers. It is also important to develop algorithms to reduce biases by recommending relevant work to researchers based on the \emph{content} of the research, independent of the popularity of the authors. Tools that are able to uncover gaps in the existing literature might even lead to a more uniform allocation of researchers to topics, reducing the bias towards overpopulated areas.
\end{movedpars}

\subsection{AI-Driven Scientific Discovery: Ideation, Hypothesis Generation, and Experimentation}
\label{ax:scientific_discovery}

\begin{figure*}
  \centering
  \includegraphics[width=0.95\textwidth]{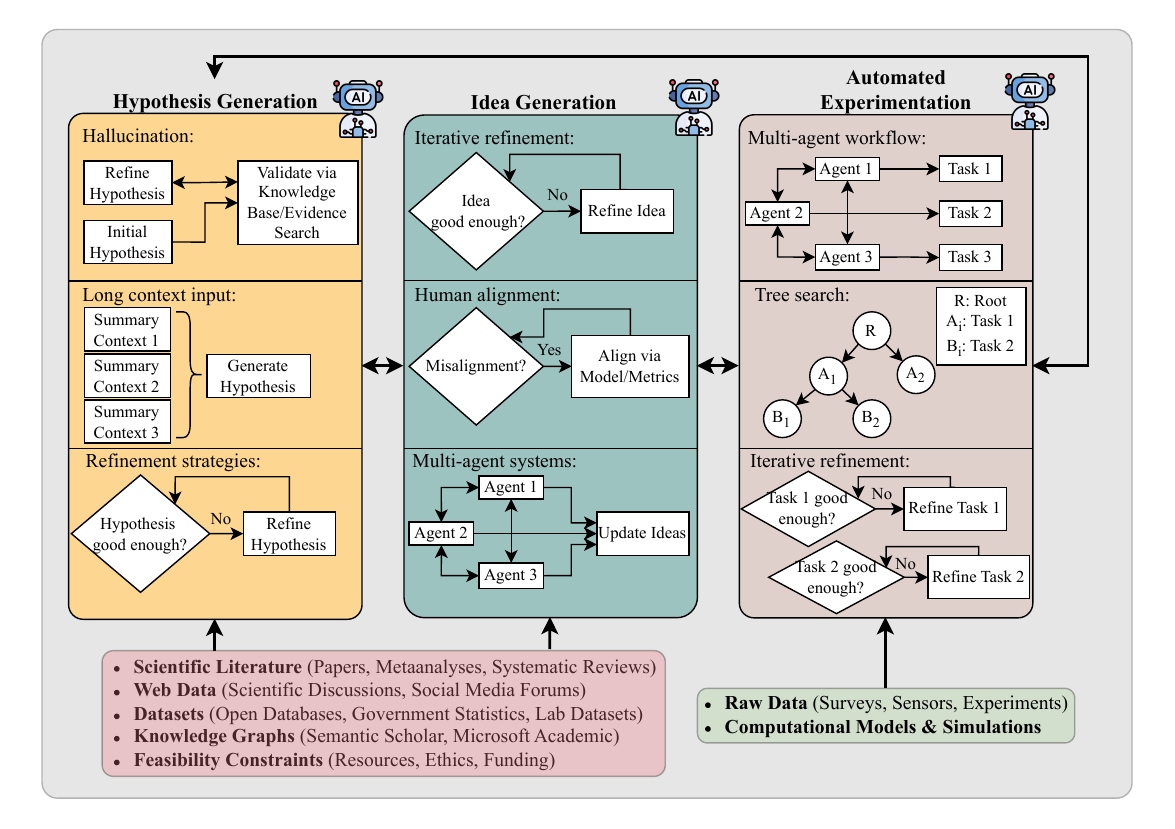} 
  \caption{Visualization of the hypothesis generation, idea generation, and automated experimentation process.}
  \label{fig:hypotheses_idea_experimentation_overview}
\end{figure*}

\paragraph{Methods} Figure~\ref{fig:hypotheses_idea_experimentation_overview} provides a broad overview of the methods applied in hypothesis generation, idea generation, and automated experimentation. 
Most works in hypothesis generation focus on reducing hallucinations, handling long contexts, and iteratively refining outputs. To reduce hallucinations, an initial hypothesis is validated against a knowledge base for refinement. For long-context inputs, different contexts are summarized and integrated, while refinement strategies iteratively improve the hypothesis until it meets a satisfactory level. A similar iterative refinement strategy is also applied in idea generation. Additionally, alignment strategies are employed to make generated ideas more thoughtful and feasible. In multi-agent systems, multiple agents collaborate to enhance the idea generation process. In contrast, automated experimentation often relies on tree search for selecting optimal examples, multi-agent workflows where LLMs collaborate on distinct tasks, and iterative refinement to improve task performance. While hypothesis and idea generation leverage diverse sources such as scientific literature, web data, and datasets, automated experimentation operates on predefined ideas and requires access to computational models, simulations, and raw data.

\begin{movedpars}
\paragraph{Ethical Concerns}
In the area of idea generation, there is a risk of reinforcing established research paradigms. LLMs trained on the basis of existing literature may favor popular paths and neglect underrepresented research directions. As a result, unconventional ideas may be unintentionally marginalized. For example, an AI might repeatedly suggest incremental improvements in a dominant field rather than proposing entirely new lines of research, thereby limiting the diversity of scientific thinking. 
LLM-generated hypotheses may also lack transparency, making it difficult to assess their validity or underlying assumptions, which could lead to flawed experiments. For example, an LLM might identify a statistical correlation in its training data and propose hypotheses without clearly revealing the underlying assumptions or data sources, making it difficult for researchers to verify its scientific soundness or hold anyone accountable if the hypotheses proves misleading. \changed{Additionally, LLMs that ideation and hypothesis generation systems rely on are not safe by design~\cite{wei2023jailbroken}. These systems may be jailbroken by malicious users to produce harmful ideas---for instance, re-purposing open science artefacts for malicious ends~\cite{hashemi2026}, and suggesting toxic molecular designs~\cite{he2023control}. \citet{zhou2026benchmarking} show that these systems may suggest unsafe experimental procedures (e.g., improper equipment use, unsafe chemical handling, or failure to recognize experimental hazards), which is problematic without human oversight.}
\end{movedpars}

\subsection{Text-based Content Generation} 
\label{ax:content_generation}

\begin{figure*}
  \centering
  \includegraphics[width=0.95\textwidth]{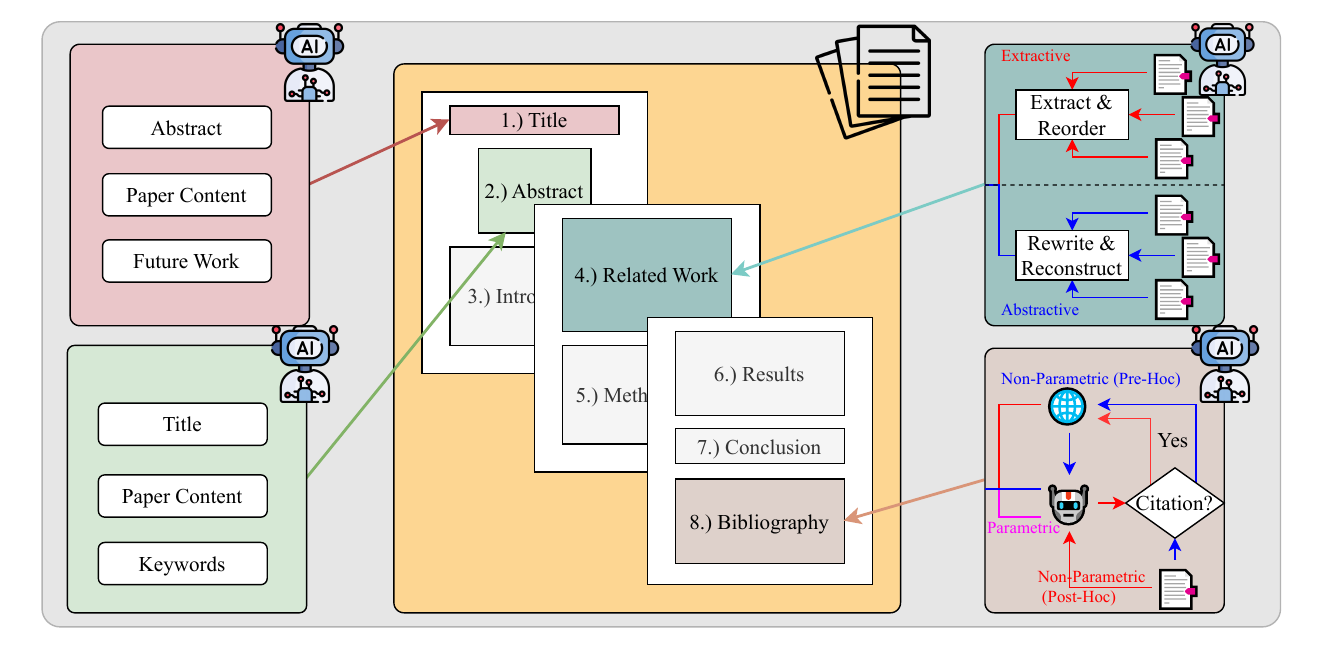} 
  \caption{Visualization of the content generation process for academic papers.}
  \label{fig:ax_content_generation_overview}
\end{figure*}

\paragraph{Methods} Figure~\ref{fig:ax_content_generation_overview} illustrates the content generation process for academic papers, covering title, abstract, related work, and bibliography generation, with their respective methods.  Title generation methods include abstract-to-title, content-to-title, and future work-to-title mappings. Abstract generation typically involves title-to-abstract and keywords-to-abstract techniques. Related work generation follows either extractive methods (reordering extracted sentences) or abstractive methods (rewriting content from multiple papers). Bibliography generation is categorized into non-parametric methods (retrieving references from external sources) and parametric methods (LLMs generating references from preexisting knowledge without retrieval). Non-parametric methods are further divided into pre-hoc (determining citation needs before text generation and retrieving references beforehand) and post-hoc (checking for citations after text generation and appending retrieved references as needed).

\begin{movedpars}
\paragraph{Ethical Concerns}
In scientific work, the issues of authorship and plagiarism in AI-generated texts are major concerns. In general, it is hard to distinguish between AI- and human-generated texts. 
Although there are a number of tools that purport to detect AI-generated text (e.g., \href{https://app.gptzero.me/}{GPTZero} and \href{https://hivemoderation.com/ai-generated-content-detection}{Hive}), \citet{anderson2023ai} show that they can be fooled by applying automatic paraphrasing. 
Studies have also found that ChatGPT-generated texts easily pass automated plagiarism detectors~\cite{else2023chatgpt,altmae2023artificial}. 
\end{movedpars}

\subsection{Multimodal Content Generation and Understanding}
\label{ax:multimodal_generation}

\subsubsection{Data} Table~\ref{tab:data_text_generation} provides an overview of datasets for multimodal content generation and understanding.

\input{table/multimodality_overview_table}

\begin{movedpars}
\paragraph{Scientific Table Understanding}

Table understanding often comes 
as table-to-text generation, which
focuses on producing accurate 
textual descriptions that reflect 
table content. 
\textbf{SciGen}~\cite{BENCHMARKS2021_149e9677} and \textbf{numericNLG}~\cite{suadaa-etal-2021-towards} are benchmarks 
specifically focused on scientific table reasoning, both emphasizing arithmetic reasoning over numerical tables. 
Each dataset contains
1.3K expert-annotated tables. 
The annotations include the tables and parts of the scientific papers that describe the corresponding findings of the annotated tables. 
A specific subtask of these benchmarks is explored in 
\citet{ampomah-etal-2022-generating}, which focuses on generating textual explanations for tables reporting ML model performance metrics. This dataset pairs numerical tables of classification performance (e.g., precision, recall, and accuracy) with expert-written textual explanations that analyze and interpret the metrics. Datasets like 
\textbf{HiTab}~\cite{cheng-etal-2022-hitab} 
tackle the complexity of hierarchical tables commonly found in statistical reports, introducing numerical reasoning tasks that require models to account for implicit relationships and hierarchical indexing within tables.
\textbf{SciXGen}~\cite{chen-etal-2021-scixgen-scientific} broadens the scope of table-to-text generation with context-aware scientific text generation. By drawing from over 200K
scientific papers, SciXGen requires models to generate descriptions for tables, figures, and algorithms, grounded in the surrounding body text.  \changed{Recent work further shows that scientific table performance depends strongly on the table \emph{source\slash modality} (e.g., PDF-rendered images vs.\ \LaTeX\slash HTML tables), and introduces dedicated multimodal scientific table benchmarks to evaluate and improve numerical reasoning~\cite{yang2025does}.}

\paragraph{Scientific Table Generation}

Table generation often comes in the form of text-to-table generation~\cite{shi2024ct,deng-etal-2024-text,jiang-etal-2024-tkgt},
the process of converting unstructured textual information into structured tabular formats.  
This process is particularly valuable for scientific domains where textual data often contains detailed experimental results, observations, or findings that need transformation into structured tables. 
In the scientific domain, \textbf{ArXivDIGESTables}~\cite{newman-etal-2024-arxivdigestables} addresses the specific challenge of automating the creation of literature review tables. Rows in these tables represent individual papers, while columns capture comparative aspects such as methods, datasets, and results. ArXivDIGESTables
supports the generation of literature review tables by leveraging additional grounding context, such as captions and in-text references. 
\end{movedpars}

\subsubsection{Methods and Results} 
\begin{figure*}
  \centering
  \includegraphics[width=0.95\textwidth]{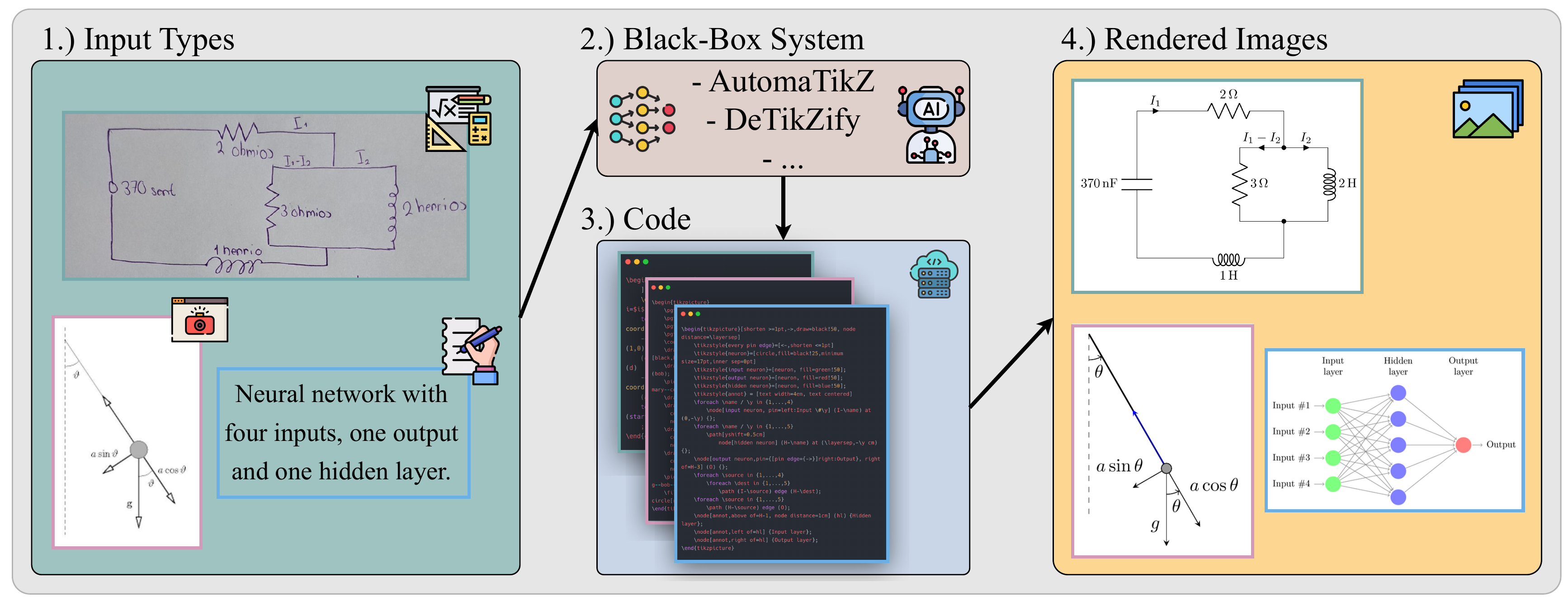} 
  \caption{Overview of the scientific figure generation process. Various input types including sketches, screenshots, and text can be used to generate TikZ code with tools such as AutomaTikZ~\cite{belouadi2024automatikz} and DeTikZify~\cite{belouadi2024detikzify}. The generated code is then rendered into high-quality vector graphics images.}
  \label{fig:figure_generation_overview}
\end{figure*}

Table~\ref{tab:section4.4_Method} provides a summary of representative approaches for multimodal content generation and understanding methods, and Fig.~\ref{fig:figure_generation_overview} illustrates the process of scientific figure generation.  The remainder of this section presents the methods for the tasks of table understanding and generation introduced above, and extends the discussion of scientific slide and poster generation methods from §\ref{sec:multimodal}.

\input{table/multimodal_table2}

\paragraph{Scientific Slide and Poster Generation}

For scientific slide generation, early works 
typically relied on heuristic rule-based approaches. For instance, \citet{Sravanthi2009SlidesGenAG} develop a rule-based system to generate slides for each section and subsection of a paper, with the textual content of the slides coming from a query-based extractive summarization system. 
Later, researchers began to leverage machine learning approaches to extract key phrases and their corresponding important sentences. \citet{Hu2013PPSGenLT} use a support vector regression~(SVR) model to learn the importance of each sentence in a paper. The slides are then generated using an integer linear programming~(ILP) model to select and align key phrases and sentences. \citet{Wang2017PhraseBasedPS} propose a system to generate slides for each section of a given paper, focusing on creating two-layer bullet points. The authors first extract key phrases from the paper using 
a parser 
and then use a random forest classifier to predict the hierarchical relationships between pairs of phrases. 
\citet{Li2021TowardsTS} develop two sentence extractors—a neural-based model and a log-linear model—within a mutual learning framework to extract relevant sentences from papers. These sentences are used to generate draft slides for four topics: contribution, dataset, baseline, and future work. 

It is important to note that all the aforementioned works focus on extracting sentences or phrases from the given paper to serve as the slide text content. In contrast, \citet{Fu2021DOC2PPTAP} and \citet{sun-etal-2021-d2s} take a different approach by training sequence-to-sequence models to generate sentences 
for the slide text content. This distinction is analogous to the difference between extractive and abstractive summaries in text summarization. More specifically, \citet{Fu2021DOC2PPTAP} design a hierarchical recurrent sequence-to-sequence architecture to encode the input document, including sentences and images, and generate a slide deck. In contrast, \citet{sun-etal-2021-d2s} assume that slide titles would be provided by end users, and use these titles to retrieve relevant and engaging text, figures, and tables from a given paper using a dense retrieval model. They then summarize the retrieved content into bullet points with a fine-tuned long-form question answering system based on BART. 

With recent advancements in 
LLMs and vision--language models~(VLMs), researchers have started using these technologies for generating scientific presentation slides. \citet{mondal-etal-2024-presentations} propose a system to generate persona-aware presentation slides by fine-tuning LLMs such as text-davinci-003 and gpt-3.5-turbo with a small training dataset containing personalized slide decks for each paper. \citet{maheshwari-etal-2024-presentations}, focusing solely on generating text content, develop an approach that combines graph neural networks (GNNs) with LLMs to capture non-linearity in presentation generation, attributing source paragraphs to each generated slide within the presentation. \citet{bandyopadhyay-etal-2024-enhancing-presentation1} design a bird's-eye view document representation to generate an outline, map slides to sections, and then create textual content for each slide individually using LLMs. The approach then extracts images from the original papers by identifying text--image similarity in a shared subspace through a VLM.

Generating posters from scientific papers has received less attention compared to scientific slide generation. \citet{Qiang2016LearningTG} introduce a graphical model to infer key content, panel layouts, and the attributes of each panel from data. The poster generator demo system of \citet{Xu2022PosterBotAS} first identifies important sections of a paper using a trained classifier, then employs a summarization model to extract key sentences and related graphs from each section to construct corresponding panels. Finally, the system generates a \LaTeX\ document for the poster based on a template selected by the user.

\begin{movedpars}
\paragraph{Scientific Table Understanding}

Table-to-text generation encompasses a range of methodologies designed to transform structured tabular data into coherent and accurate textual descriptions. These techniques process, reason over, and utilize tabular structures to address challenges such as logical reasoning, content fidelity, and domain-specific adaptation. 
\emph{Serialization} is a foundational approach where tables are linearized into sequences compatible with transformer-based language models. In this method, tables are converted into linear text sequences using special characters to delineate structure~\cite{BENCHMARKS2021_149e9677,parikh-etal-2020-totto,andrejczuk-etal-2022-table}. 
\emph{Structure-aware methods} explicitly model the inherent relationships and hierarchies within tables to enhance reasoning and generation fidelity. These include \emph{intermediate representations}~\cite{zhao-etal-2023-loft,zhao-etal-2023-sortie,10.1162/tacl_a_00641}, 
\emph{structure-aware pretraining}~\cite{petrak-etal-2023-arithmetic,korkmaz-del-rio-chanona-2024-integrating,10707812}, and \emph{structure-aware self-attention mechanisms}~\cite{wang-etal-2022-robust,10.1145/3622896.3622919}. 
For \textbf{evaluation}, common (if flawed) metrics like BLEU and BARTScore are widely used to evaluate the fluency and relevance of generated text against reference outputs. However, ensuring faithfulness to the source table remains a significant challenge, often requiring human evaluation for accurate assessment~\cite{BENCHMARKS2021_149e9677,petrak-etal-2023-arithmetic}.

\paragraph{Scientific Table Generation} 

While no existing approach to table generation focuses specifically on scientific 
data, several methodologies present promising directions. 
The gTBLS (Generative Tables) approach~\cite{sundar2024gtbls} proposes a two-stage table generation process. The first stage infers the table structure from input text, while the second stage generates table content by formulating table-guided questions; 
this enhances
syntactic validity and logical coherence of generated tables.
In the context of open-structure table extraction, OpenTE~\cite{10448427} tackles the task of extracting tables with intrinsic semantic, calculational, and hierarchical structure from unstructured text. OpenTE introduces 
a three-step pipeline that identifies semantic and relational connections among table columns, extracts structured data, and grounds the output by aligning extracted data with the source text and table structure.
\textbf{Evaluation} of text-to-table generation for science should focus on structural accuracy, value fidelity, and semantic coherence. TabEval~\cite{ramu-etal-2024-bad} provides a promising direction by introducing a decomposition-based framework that breaks tables into atomic statements and evaluates them using entailment-based measures, though comprehensive evaluation still requires further advancements.

\subsubsection{Ethical Concerns}
Tools for figure, table, slide, and poster generation are technically limited \changed{by the relatively small sizes of datasets for training and testing. For example, AutomaTikZ~\cite{belouadi2024automatikz} and its extensions contain only several hundred thousand pairs of textual descriptions and corresponding code snippets, while general-purpose image generation datasets are often orders of magnitudes larger. There is also a misalignment problem of image captions and the corresponding images\slash code~\cite{greisinger2026tikzilla}, increasing the risk of hallucinations.  These tools can therefore easily produce incorrect scientific figures, particularly when their users overlook, ignore, or maliciously abuse their limitations.}
\end{movedpars}

\begin{movedpars}
\subsection{Peer Review}
\label{ax:peerreview}

\paragraph{Assessment of Scientific Rigor} 

Several attempts have been made to computationally analyze the rigor of scientific papers. \citet{Wael}, for example, investigated how researchers use the word ``rigor'' in information system literature and discovered that the exact meaning was ambiguous in current research. Nonetheless, various automated tools have been proposed to assess the rigor of academic papers. \citet{phillips2017online} develop an online software that spots genetic errors in cancer papers, and \citet{sun2022assessing} use knowledge graphs to assess the credibility of papers based on metadata such as publication venue, affiliation, and citations. 
However, these methods are neither domain-specific, nor do they provide sufficient guidance for authors to improve their narrative and writing.
In contrast, SciScore~\cite{SciScore_2024} uses language models to produce rigor reports for paper drafts with the aim of helping authors identify weaknesses in their presentation. 
More recently, \citet{james-etal-2024-rigour} propose a bottom-up, data-driven framework that automates the identification and definition of rigor criteria while assessing their relevance in scientific texts. Their framework integrates three key components: rigor keyword extraction, detailed definition generation, and the identification of salient criteria. Additionally, its domain-agnostic design allows for flexible adaptation across different fields.

\paragraph{Scientific Claim Verification} 

The increasing volume of scientific literature has created a demand for automated methods for verifying the validity and reliability of research claims. Scientific fact verification, which aims to assess the accuracy of scientific statements, often relies on external knowledge to support or refute claims~\cite{vladika2023scientific, dmonte2024claim}. Several datasets have been developed to address this, including SciFact-Open~\cite{wadden2022scifact}, which provides scientific claims and supporting evidence from abstracts. However, it is limited to the use of abstracts as the primary source of evidence. As the statements in abstract can also be inaccurate or misleading, it is important to corroborate them with evidence from the main body of the paper. To this end, \citet{glockner-etal-2024-missci,glockner2024groundingfallaciesmisrepresentingscientific} propose a theoretical argumentation model to reconstruct fallacious reasoning of false claims that misrepresent scientific publications.
The need to contextualize claims with supporting evidence is also highlighted by \citet{chan2024overview}, who introduce a dataset of claims extracted from lab notes. Unlike other datasets, this resource is claimed to be ``actually in use'', providing a more realistic understanding of how researchers interact with scientific findings. The authors annotate claims with links to figures, tables, and methodological details, and develop associated tasks to improve retrieval. While this provides valuable resources for context-based verification, it primarily focuses on factual verification and does not evaluate the potential for overstatement.
Beyond factual correctness, there is a growing recognition for the need to analyze how researchers present their findings, rather than their mere factuality. This includes the detection of overstatements, where authors exaggerate their achievements, and understatements, where the true impact of the research is downplayed~\cite{kao2024we}.   \citet{schlichtkrull2023intended} present a qualitative analysis of how intended uses of fact verification are described in highly-cited NLP papers, particularly focusing on the introductions of the papers, to understand how these elements are framed. The work suggests that claims should be supported by relevant prior work and empirical results.

\paragraph{Ethical Concerns}
Given the critical role of scientific peer review for science, and, accordingly, for society as a whole, ethical considerations around AI-supported peer review are of utmost importance. As the general concerns around unfair biases in AI and the resulting harms apply~\cite{kuznetsov2024can}, research on safe peer-reviewing support needs to be prioritized. For instance, \citet{10.1001/jama.2023.24641} recently showed that 
LLMs exhibit 
affiliation biases when reviewing abstracts. In this context, any AI support for peer reviewing needs to be critically evaluated~\cite{schintler2023critical}, and solutions that target only a particular aspect in a collaborative environment that leaves the scientific autonomy to the human expert may be preferable to end-to-end reviewing systems. \changed{A recent discussion on specific risks of using LLMs in peer reviewing is also provided by \citet{BenSaad2025-BENTAT-27}, which highlights the risk of diminished engagement and critical thinking of reviewers and call for the creation of ethical standards to balance AI's capabilities with human expertise.}
\end{movedpars}

\section{This Paper as an AI Use Case}
\label{ax:ai_usage}

The preparation of this survey paper itself involved the use of AI tools to support specific aspects of the research workflow. For retrieving, selecting, and categorizing the literature and resources described in the various task subsections of §\ref{sec:tasks}, many of us relied not only on traditional information retrieval tools such as Google Search and Google Scholar, but also on tools incorporating generative AI, such as NotebookLM, ChatGPT, and Scholar Inbox.  LLMs also assisted some co-authors with grammar and spell checking, as well as generating \LaTeX\ code for formatting tables.


%% file: table/multimodality_overview_table.tex
\begin{table}
\begin{changedfloat}
\caption{Multimodal content generation and understanding datasets.}
\label{tab:section4.4_data}
\begin{small}
    \begin{tabular}{p{3cm}|p{6cm}|p{5cm}} 
        \toprule
      \textbf{Dataset} & \textbf{Size} & \textbf{Data Sources}  
      \\
    \midrule  
     \multicolumn{3}{c}{\textbf{Scientific Figure Understanding}} \\ \hline
arXivCap~\cite{li-etal-2024-multimodal-arxiv} & 6.4M images and 3.9M captions from 572K papers & arXiv \\ 
FigureQA~\cite{ebrahimi2018figureqa} & $>$ 100K scientific-style figures& Synthetic\\
ChartQA~\cite{masry-etal-2022-chartqa} & 4.8K charts, 9.6K QA pairs& statista.com, pewresearch.com, etc. \\ 
CharXiv~\cite{wang2024charxiv} & 2.3K charts with descriptive and reasoning questions& arXiv\\
arXivQA~\cite{li-etal-2024-multimodal-arxiv} & 35K figures with 100K QA pairs& arXiv\\
SPIQA~\cite{pramanick2024spiqa} & 152K figures with 270K QAs & 19 top-tier academic conferences\\
ChartSumm~\cite{xu2024chartadapterlargevisionlanguagemodel} & 84K charts & Knoema\\ 
SciMMIR~\cite{wu-etal-2024-scimmir} & 530K  figures and tables image--text pairs & arXiv \\
\hline
 \multicolumn{3}{c}{\textbf{Scientific Figure Generation}} \\ \hline
{DaTikZ (V1-V3)~\cite{belouadi2024automatikz,belouadi2024detikzify,belouadi2025tikzero}} & {118–456K pairs of captions/TikZ code} & {arXiv, \TeX\ Stack Exchange} \\
{DaTikZ V4~\cite{greisinger2026tikzilla}} & {2M pairs of VLM descriptions\slash TikZ code} & {arXiv, GitHub, \TeX\ Stack Exchange} \\
{DiagramGenBench~\cite{wei2025words}} & {6713 train\slash 270 test for coding/generation, 1400 train\slash 200 test for editing} & {VGQA and DaTikZ licensed under CC BY 4.0 or MIT} \\
{Plot2XML~\cite{cui2025draw}} & {247 complex diagrams} & {Conference papers} \\
{PandasPlotBench~\cite{galimzyanov2025drawing}} & {175 visualizations} & {Matplotlib gallery} \\
{VisCode-200K~\cite{ni2025viscoder}} & {200K supervised examples} & {Open-source Python repositories, Code--Feedback dataset~\cite{zheng2024opencodeinterpreter}} \\
ScImage~\cite{zhang2024scimagegoodmultimodallarge} & 404 instructions and 3K generated scientific images & Manual (template) construction \\
SciDoc2DiagramBench \cite{mondal-etal-2024-scidoc2diagrammer} & 1,080 extrapolated diagrams in the format ``<paper(s), intent of diagram, gold diagram>'' & ACL Anthology \\ 
ChartMimic~\cite{shi2024chartmimicevaluatinglmmscrossmodal} & 1000 triplets of (figure, instruction, code) instances & Physics, Computer Science, Economics, etc.\\
\hline
\multicolumn{3}{c}{\textbf{Scientific Table Understanding}} \\ \hline
SciGen~\cite{BENCHMARKS2021_149e9677} & 1.3K pairs of scientific tables and their descriptions & arXiv (especially cs.CL and cs.LG) \\ 
 NumericNLG~\cite{suadaa-etal-2021-towards} & 1.3K pairs of scientific tables and their descriptions & ACL Anthology \\
SciXGen~\cite{chen-etal-2021-scixgen-scientific} & 484K tables from 205K papers & arXiv  \\
\hline 
    \multicolumn{3}{c}{\textbf{Scientific Table Generation}} \\ \hline
arXivDigestTables~\cite{newman-etal-2024-arxivdigestables} & 2,228 literature review tables extracted from arXiv papers that synthesize a total of 7,542 research paper &  literature review tables from arXiv papers from April 2007 to
November 2023 
\\
\hline
  \multicolumn{3}{c}{\textbf{Scientific Slides and Poster Generation}} \\ \hline
SciDuet~\cite{sun-etal-2021-d2s} & 1,088 papers and 10,034 slides by their authors & NeurIPS/ICML/ACL  Anthology  \\ 
DOC2PPT~\cite{Fu2021DOC2PPTAP} & 5,873 papers and 98,856 slides by their authors  & CV (CVPR, ECCV,
BMVC), NLP (ACL, NAACL, EMNLP), ML (ICML, NeurIPS, ICLR) \\ 
Persona-Aware-D2S~\cite{mondal-etal-2024-presentations} & 75 papers from SciDuet, and 300 slides 
& ACL Anthology \\
{SlidesBench~\cite{ge2025autopresent}} & {7K training and 585 test}
& {Web (Art, Marketing, Environment, Technology, etc.)}\\
\bottomrule
\end{tabular}
\end{small}
\end{changedfloat}
\end{table}


%% file: table/multimodal_table2.tex
\begin{table}
\caption{Multimodal content generation and understanding approaches.}
\label{tab:section4.4_Method}
\centering
\begin{small}
    \begin{tabular}{p{2.5cm}|p{2.5cm}|p{2cm}|p{1.7cm}|p{1.8cm}|p{2cm}}
        \toprule
        \textbf{Task} & \textbf{Input} & \textbf{Output} & \textbf{Dataset} & \textbf{Method} & \textbf{Evaluation} \\
        \midrule  
        \multicolumn{6}{c}{\textbf{Scientific Figure Understanding}} \\ 
        \midrule
        Question Answering \cite{10.1007/978-3-319-46493-0_15} & Synthetic, scientific-style figures and questions & Answers & FigureQA & Fine-tuning & Accuracy \\ 
        Chart Summarization \cite{Rahman2023ChartSummAC} & Chart images with metadata & Chart summaries & ChartSumm & Fine-tuning & Automatic evaluation \\ 
        Caption Figure Retrieval & Figure or caption & Caption or figure & SciMMIR & Fine-tuning & Ranking Metrics \\
        \midrule
        \multicolumn{6}{c}{\textbf{Scientific Figure Generation}} \\ 
        \midrule
        Caption/Instruction-to-code generation \citep{belouadi2024automatikz,voigt2024plots} & (Extended) scientific caption or instruction & Compilable (TikZ, Vega, etc.) code of scientific figure & AutomaTikZ, DaTikZ & Fine-tuning & Human \& various metrics \citep{belouadi2024automatikz} \\
        Description-to-image generation \citep{zhang2024scimagegoodmultimodallarge} & Description/instruction & Scientific image & ScImage & Prompting & Human \\
        Sketch/Image-to-image generation & Scientific (raster) image or sketch & Compilable TikZ code of scientific figure & DaTikZ-v2 & Fine-tuning \& MCTS & Human \& various metrics \\
        Scientific diagram generation \cite{mondal-etal-2024-scidoc2diagrammer} & Scientific paper(s) + intent & Diagram & SciDoc2-DiagramBench & Two-stage pipeline & Human \& various metrics \\ 
        \midrule
        \multicolumn{6}{c}{\textbf{Scientific Table Understanding}} \\ 
        \midrule
        Table description \cite{BENCHMARKS2021_149e9677} & Tables from scientific articles & Table description & SciGen & Fine-tuning & Automatic \& human evaluation \\
        Numerical reasoning \cite{suadaa-etal-2021-towards} & Tables from scientific papers & Numerical descriptions & NumericNLG & Fine-tuning & Automatic \& human evaluation \\  
        \midrule
        \multicolumn{6}{c}{\textbf{Scientific Table Generation}} \\ 
        \midrule
        Literature review table generation \cite{newman-etal-2024-arxivdigestables} & A list of papers & Table schema + values & ArXivDigest Tables & Prompting & Automatic \& human evaluation \\
        \midrule
        \multicolumn{6}{c}{\textbf{Scientific Slide and Poster Generation}} \\ 
        \midrule
        Single slide generation \cite{sun-etal-2021-d2s} & Paper + slide title & Slide content & SciDuet & Two-step method & ROUGE and human evaluation \\ 
        Slide deck generation \cite{Fu2021DOC2PPTAP} & Paper & A deck of slides & DOC2PPT & Hierarchical generative model & Automatic \& human evaluation \\ 
        Personalized slide deck generation \cite{mondal-etal-2024-presentations} & Paper + target audience (technical or non-technical) & A deck of slides & Persona-Aware-D2S & Fine-tuning & Automatic \& human evaluation \\
        \bottomrule
    \end{tabular}
\end{small}
\end{table}
